%% file: icml.tex
\documentclass{article}

\usepackage{microtype}
\usepackage{graphicx}
\usepackage{subcaption}
\usepackage{booktabs} 

\usepackage{hyperref}
\usepackage{xspace}       
\usepackage{xspace}       
\usepackage{colortbl}     
\usepackage{multirow}
\usepackage{adjustbox}
\usepackage{makecell}

\usepackage[accepted]{icml2026}

\usepackage{amsmath}
\usepackage{amssymb}
\usepackage{mathtools}
\usepackage{amsthm}

\usepackage[capitalize,noabbrev]{cleveref}

\theoremstyle{plain}

\theoremstyle{definition}

\theoremstyle{remark}

\usepackage[textsize=tiny]{todonotes}
\usepackage{tcolorbox}

\icmltitlerunning{\methodname: Bayesian Decomposition of Vision Language Action Models via Latent Action Queries}

\begin{document}

\newcommand{\methodname}{\text{LangForce}\xspace} 

\newcommand{\infobox}[1]{
    \begin{tcolorbox}[
        colback=white!90!gray,     
        colframe=cyan!40!black,   
        arc=5pt,                   
        boxsep=5pt,                 
        left=5pt,                  
        right=10pt,                 
        top=2pt,                   
        bottom=3pt,                
        boxrule=0.8pt,                      
    ]
    \vspace{-0.1cm}
         \textit{#1}
    \vspace{-0.1cm}
    \end{tcolorbox}
}

\twocolumn[
  \icmltitle{\methodname: Bayesian Decomposition of Vision Language Action Models \\via Latent Action Queries}



  \icmlsetsymbol{equal}{*}
  \makeatletter
  \newcounter{@affilHUST}\setcounter{@affilHUST}{1}
  \newcounter{@affilZGCA}\setcounter{@affilZGCA}{2}
  \newcounter{@affilZGCI}\setcounter{@affilZGCI}{3}
  \newcounter{@affilHIT}\setcounter{@affilHIT}{4}
  \newcounter{@affilHKUSTGZ}\setcounter{@affilHKUSTGZ}{5}
  \newcounter{@affilZZU}\setcounter{@affilZZU}{6}
  \newcounter{@affilBUAA}\setcounter{@affilBUAA}{7}
  \newcounter{@affilECNU}\setcounter{@affilECNU}{8}
  \newcounter{@affilDeepCybo}\setcounter{@affilDeepCybo}{9}
  \setcounter{@affiliationcounter}{9}
  \makeatother

  \begin{icmlauthorlist}
    \icmlauthor{Shijie Lian}{equal,HUST,ZGCA}
    \icmlauthor{Bin Yu}{equal,ZGCA,HIT}
    \icmlauthor{Xiaopeng Lin}{equal,ZGCI,HKUSTGZ}
    \icmlauthor{Laurence Tianruo Yang}{ZZU,HUST}
    \icmlauthor{Zhaolong Shen}{ZGCA,BUAA}
    \icmlauthor{Changti Wu}{ZGCA,ECNU}
    \icmlauthor{Yuzhuo Miao}{ZGCA,HIT}
    \icmlauthor{Cong Huang}{ZGCA,ZGCI}
    \icmlauthor{Kai Chen}{ZGCA,ZGCI,DeepCybo}
  \end{icmlauthorlist}

  \icmlaffiliation{HUST}{Huazhong University of Science and Technology}
  \icmlaffiliation{ZGCA}{Beijing Zhongguancun Academy}
  \icmlaffiliation{ZGCI}{Zhongguancun Institute of Artificial Intelligence}
  \icmlaffiliation{HIT}{Harbin Institute of Technology}
  \icmlaffiliation{HKUSTGZ}{The Hong Kong University of Science and Technology (Guangzhou)}
  \icmlaffiliation{ZZU}{Zhengzhou University}
  \icmlaffiliation{BUAA}{Beihang University}
  \icmlaffiliation{ECNU}{East China Normal University}
  \icmlaffiliation{DeepCybo}{DeepCybot Co., Ltd.}

  \icmlcorrespondingauthor{Laurence Tianruo Yang}{ltyang@ieee.org}
  \icmlcorrespondingauthor{Kai Chen}{kaichen@zgci.ac.cn}

  \icmlkeywords{Machine Learning, ICML}

  \vskip 0.3in
]



\printAffiliationsAndNotice{Work done at Beijing Zhongguancun Academy.\\\hspace*{1.4\footnotesep}\textsuperscript{*}Equal contribution. }

\input{sec/0_abstract}
\input{sec/1_intro}
\input{sec/2_motivation}
\input{sec/3_method}
\input{sec/4_experiment}
\input{sec/6_related}

\input{sec/7_conclusion}




\section*{Acknowledgements}

This work is supported in part by the National Natural Science Foundation of China under Grant U23A20300; in part by the Beijing Zhongguancun Academy (Grant No. C20250510); in part by the High Innovation Plan (Grant No. 202504841022).

\section*{Impact Statement}
This paper presents work whose goal is to advance the field of Machine Learning. There are many potential societal consequences of our work, none which we feel must be specifically highlighted here.

\bibliography{custom}
\bibliographystyle{icml2026}

\newpage
\appendix
\onecolumn

\input{sec/x_appendix}

\end{document}

%% file: sec/0_abstract.tex
\begin{abstract}
Vision-Language-Action (VLA) models have shown promise in robot manipulation but often struggle to generalize to new instructions or complex multi-task scenarios.
We identify a critical pathology in current training paradigms where goal-driven data collection creates a dataset bias.
In such datasets, language instructions are highly predictable from visual observations alone, causing the conditional mutual information between instructions and actions to vanish, a phenomenon we term \textit{Information Collapse}.
Consequently, models degenerate into vision-only policies that ignore language constraints.
To address this, we propose \textbf{\methodname}, enforces instruction following via Bayesian decomposition.
By introducing learnable \textbf{Latent Action Queries}, we construct a dual-branch architecture to estimate both a vision-only prior $p(a \mid v)$ and a language-conditioned posterior $\pi(a \mid v, \ell)$.
We then optimize the policy to maximize the conditional Pointwise Mutual Information (PMI) between actions and instructions.
This objective effectively penalizes the vision shortcut and rewards actions that explicitly explain the language command.
Extensive experiments across on three benchmarks demonstrate substantial gains, including an \textbf{11.3\%} improvement on the challenging OOD SimplerEnv benchmark, validating the ability of \methodname to robustly ground language in action.
Code and videos are available at \href{https://github.com/ZGC-EmbodyAI/LangForce}{this}.
\end{abstract}


%% file: sec/1_intro.tex
\section{Introduction}
\label{sec:intro}

Vision-Language-Action (VLA) models \citep{OpenVLA_2024_CoRL, RDT-1B_2025_ICLR, GR00T_2025_arXiv, PI05_2025_arXiv} have emerged as a promising paradigm for general-purpose robot manipulation, leveraging the vast knowledge of pre-trained Vision-Language Models (VLMs) to ground natural language instructions into physical actions.
By training on large-scale datasets of human demonstrations, these models aim to learn a policy $\pi(a \mid v, \ell)$ that can execute diverse tasks specified by language $\ell$ given visual observations $v$.

While demonstrating strong performance in in-distribution settings, current VLA models still face challenges in generalizing to novel instructions or complex multi-task scenarios, particularly in out-of-distribution (OOD) environments \citep{xu2025seeing, xing2025shortcut, xu2023joint}. This limitation is especially pronounced during post-training, where fine-tuning on narrow, task-specific datasets can lead to catastrophic forgetting of the VLM's general capabilities and impair its ability to generalize to new tasks.
We hypothesize that this fragility is exacerbated by a prevalent bias in current robotic datasets.
Most robotic datasets are collected in a goal-driven manner, where a human operator performs a specific task repeatedly in a fixed scene. In such datasets, the mapping from visual scene $v$ to language instruction $\ell$ is nearly injective: seeing a cabinet in the scene almost invariably implies the task ``open the cabinet,'' while seeing a bottle implies ``pick up the bottle.'' This deterministic coupling results in a sharp conditional distribution $p(\ell \mid v)$.

\begin{figure*}[!th]
    \centering
    \includegraphics[width=1\textwidth]{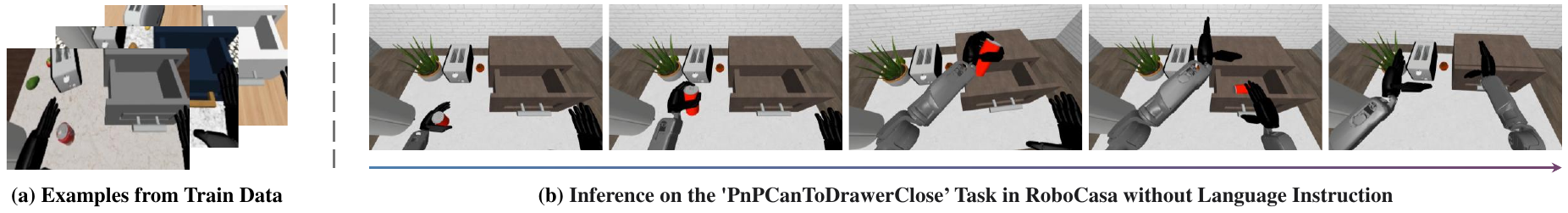}
    \caption{\textbf{Examples of the vision shortcut in RoboCasa \citep{RoboCasa_2024_RSS}}. Training data exhibits visual diversity but limited task diversity. As a result, the model learns to execute tasks directly based on specific visual cues rather than relying on language instructions.}
    \label{fig:robocasa_sample}
\end{figure*}

From a Bayesian perspective, the optimal policy can be decomposed as:
\begin{equation}
    \label{eq:bayesian}
    \pi(a \mid v, \ell) = \frac{p(\ell \mid a, v) \, p(a \mid v)}{p(\ell \mid v)}.
\end{equation}
Here, $p(a \mid v)$ represents a vision-only prior (i.e., what actions are likely in this scene?), and $p(\ell \mid a, v)$ is the likelihood (i.e., how well does action $a$ explain instruction $\ell$?). 
When $p(\ell \mid v)$ is sharp, the model can predict $\ell$ solely from $v$ without attending to $a$. 
Consequently, the likelihood term $p(\ell \mid a, v)$ collapses to $p(\ell \mid v)$, and the posterior policy degenerates to the prior: 
\begin{equation}
    \label{eq:vision_shortcut}
    \pi(a \mid v, \ell) \approx p(a \mid v).
\end{equation}
In other words, the model effectively ignores the language instruction, learning a ``vision shortcut'' that fails whenever the task is ambiguous or the environment changes. 

To address this, we propose \textbf{\methodname}, a novel framework that explicitly enforces instruction following via Bayesian decomposition.
Our key insight is to maximize the conditional Pointwise Mutual Information (PMI) between actions and instructions, which is equivalent to maximizing the log-likelihood ratio (LLR): $\log p(\ell \mid a, v) - \log p(\ell \mid v)$.
This objective penalizes the vision shortcut by requiring the action $a$ to provide \textit{additional} information about $\ell$ that cannot be inferred from $v$ alone.

We instantiate this framework by introducing \textbf{Latent Action Queries}---a set of learnable tokens injected into the VLM.
These queries serve a dual purpose: they act as a bottleneck to extract action-relevant features for a downstream Diffusion Transformer (DiT) policy, and they enable a dual-branch training strategy.
In the \textit{Priori Branch}, queries attend only to vision to learn $p(a \mid v)$; in the \textit{Posteriori Branch}, they attend to both vision and language to learn $\pi(a \mid v, \ell)$. By optimizing the LLR between these branches, \methodname learns to ground language robustly without requiring new data.
Our contributions are three-fold:
\begin{enumerate}
    \item We identify and empirically validate the ``vision shortcut'' pathology in current VLA training, showing that standard models often ignore language in favor of dataset-specific visual correlations.
    \item We propose \methodname, leveraging Latent Action Queries and a dual-branch Bayesian objective to recover language-conditioned policies from biased data.
    \item We demonstrate that \methodname achieves state-of-the-art performance on SimplerEnv and RoboCasa, with a remarkable \textbf{8.8\%} improvement in OOD generalization on SimplerEnv, proving its effectiveness in breaking the vision shortcut.
\end{enumerate}


%% file: sec/2_motivation.tex
\section{Motivation: Vision Shortcut}\label{sec:motivation}

Before detailing our method, we present empirical evidence to substantiate our hypothesis: that standard VLA models trained on goal-driven datasets often learn a vision-only policy $p(a \mid v)$ rather than a true language-conditioned policy $\pi(a \mid v, \ell)$. 
Specifically, we employ the Qwen3VL-4B-GR00T model from starVLA~\citep{starvla_2025} as our representative VLA architecture.
We conduct three pilot experiments to reveal this \textbf{illusion of instruction following}.
In all three experiments, we train the model by feeding \textit{only} the visual observation $v$ (masking the language instruction $\ell$), effectively testing the vision-only prior $p(a \mid v)$.

\begin{figure*}[!th]
    \centering
    \includegraphics[width=0.98\textwidth]{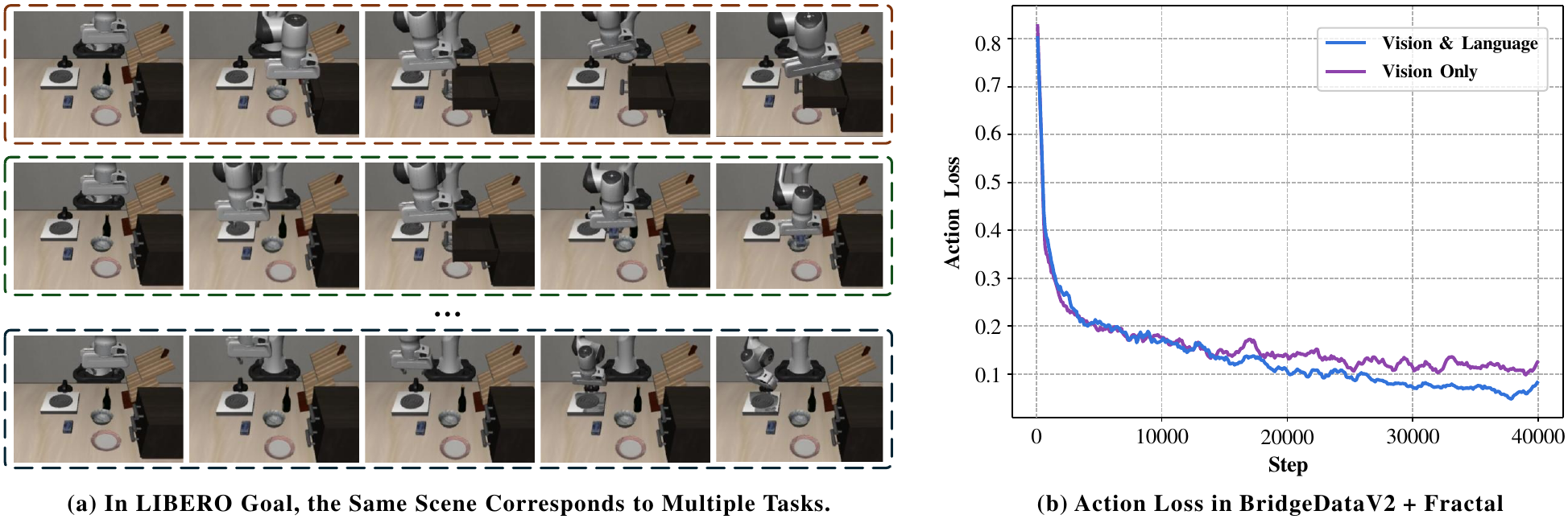}
    \caption{(a) In LIBERO Goal~\citep{LIBERO_2023_NeurIPS}, the same scene corresponds to multiple tasks, revealing ambiguity that vision-only models fail to resolve. (b) Action loss curves on BridgeDataV2~\citep{Bridgedatav2_2023_CoRL} and Fractal \citep{RT-1_2022_arXiv} show the vision-only model achieves comparable loss to the full model, indicating the presence of visual shortcuts even in diverse, in-the-wild datasets.}
    \label{fig:libero_bridge}
    \vskip -0.1in
\end{figure*}

\subsection{Experiment 1: The Vision Shortcut in ID Testing}
We first train a standard VLA model on a subset of the Humanoid robot tabletop manipulation data from PhysicalAI-Robotics-GR00T-X-Embodiment-Sim~\citep{GR00T_2025_arXiv} and evaluate on 24 tasks from the RoboCasa benchmark~\citep{RoboCasa_2024_RSS}.
Averaged across all 24 tasks, the vision-only model achieves a success rate of 44.6\%, which is close to the language-conditioned baseline of 47.8\% (as shown in Table~\ref{tab:robocasa_main_tab}). 
This small gap reveals that the model can succeed without relying on language instructions, as the training and evaluation scenes and tasks are highly similar, enabling the model to learn a near-deterministic mapping from vision to action. Figure~\ref{fig:robocasa_sample} provides a relevant example.

\subsection{Experiment 2: Failure in Ambiguous Scenarios}
To further investigate this behavior, we train a standard VLA model on the LIBERO benchmark~\citep{LIBERO_2023_NeurIPS}, which contains four subsets: Spatial, Object, Long, and Goal.
We train on all four training sets and evaluate on all four test sets.
The vision-only model achieves success rates comparable to the full VLA model on three subsets (Spatial: 90.2\%, Object: 99.6\%, Long: 86.0\% in Vision-Only, Spatial: 97.8\%, Object: 98.8\%, Long: 92.0\% in Baseline), where each visual scene corresponds to a single task.
However, on the LIBERO Goal subset, the vision-only success rate plummets to 9.8\% (97.4\% in Baseline).


The key difference is that LIBERO Goal presents inherent ambiguity: multiple valid tasks are associated with the same object configuration during training.
For instance, a scene with multiple bowls, a stove, and a drawer could correspond to either ``put bowl in drawer'' or ``put bowl on stove''.
This confirms that while the model can exploit vision-action correlations in unambiguous datasets, when multiple tasks share the same visual context, due to a lack of language to resolve ambiguity, the model is dominated by the prior $p(a \mid v)$ learned from dataset statistics.
Figure~\ref{fig:libero_bridge}(a) illustrates examples where the same visual scene in LIBERO Goal corresponds to multiple distinct tasks.

\subsection{Experiment 3: Failure in OOD Generalization}
Finally, we test the generalization capability by training on the high-quality BridgeDataV2 dataset~\citep{Bridgedatav2_2023_CoRL} and Fractal \citep{RT-1_2022_arXiv} dataset (diverse, in-the-wild scenes) and evaluating on SimplerEnv~\citep{SimplerEnv_2024_CoRL} (simulation, OOD).
During training on BridgeDataV2 and Fractal, the vision-only model achieves an action loss of 0.13, comparable to the full language-conditioned model's loss of 0.08 (as shown in Figure~\ref{fig:libero_bridge}(b)).
This indicates that even in diverse, in-the-wild scenarios, the model can still identify visual shortcuts (e.g., specific lighting or background features mapping to specific actions) to minimize the training objective without truly grounding the language instructions.
This open-loop loss trend serves as an optimization-level indication that vision-only shortcuts can fit the training distribution reasonably well.

The policy-level evidence comes from downstream execution.
When evaluated on SimplerEnv, which presents visually distinct simulation environments, the vision-only baseline achieves near 0\% success despite fitting the training distribution reasonably well.
This is the key evidence for the vision shortcut: it can produce seemingly reasonable optimization behavior on the training domain, while failing catastrophically when the policy is executed out of distribution.

\subsection{Theoretical Insight: Information Collapse}\label{subsec:information_collapse}
We formalize the ``vision shortcut'' as a collapse of the conditional mutual information (CMI) between instructions and actions. Ideally, a robust VLA policy should maintain high $I(\ell; a \mid v)$, meaning the action choice significantly reduces uncertainty about the instruction.
However, the CMI is bounded by the conditional entropy of the language: 
\begin{equation}
    I(\ell; a \mid v) = H(\ell \mid v) - H(\ell \mid a, v) \le H(\ell \mid v).
\end{equation}
In goal-driven datasets, the deterministic mapping $v \to \ell$ implies $H(\ell \mid v) \approx 0$ \citep{xu2025seeing}. Consequently, $I(\ell; a \mid v)$ is forced to vanish, theoretically preventing the model from learning any dependency between $a$ and $\ell$ beyond what is already captured by $v$.
To break this deadlock, we cannot rely on standard imitation learning. 
Instead, we must explicitly intervene to maximize the \textit{information gain} provided by the action. This motivates our use of the Log-Likelihood Ratio (LLR), which effectively estimates the Pointwise Mutual Information (PMI), rewarding the policy only when it captures the specific semantics of $\ell$ that are \textit{not} predictable from $v$.

%% file: sec/3_method.tex
\section{Method: \methodname}\label{sec:method}

\begin{figure*}[!th]
    \centering
    \includegraphics[width=0.9\textwidth]{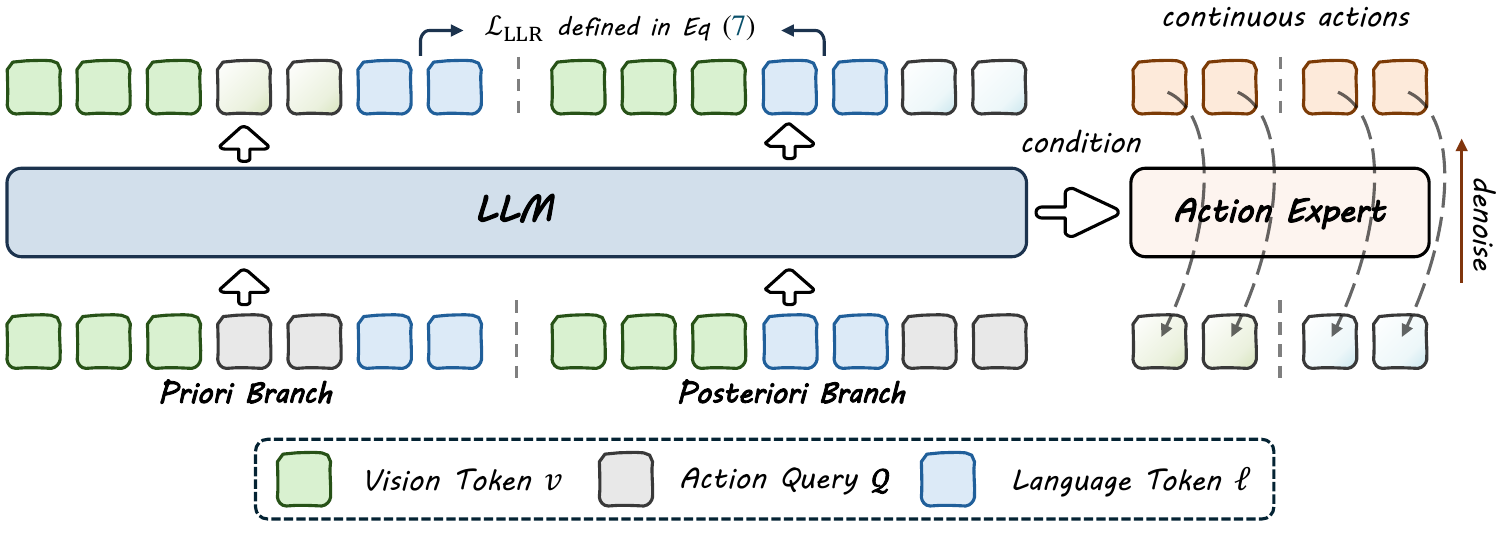}
    \caption{\textbf{The framework of \methodname}. The framework employs a dual-branch architecture with shared VLM weights. The Priori Branch (left) processes $[v, \mathcal{Q}, \ell]$ with causal masking to learn the vision-only prior $p(a \mid v)$. The Posteriori Branch (right) processes $[v, \ell, \mathcal{Q}]$ to learn the full policy $\pi(a \mid v, \ell)$. Latent Action Queries $\mathcal{Q}$ serve as a bottleneck interface, and the LLR objective (in Eq. \ref{eq:llr}) encourages the model to maximize the information between actions and instructions. At inference, only the Posteriori Branch is used, incurring no additional computational overhead.}
    \label{fig:main_framework}
    \vskip -0.1in
\end{figure*}

In this section, we introduce \textbf{\methodname}, a framework designed to mitigate the vision shortcut in VLA models. We present the overall framework of \methodname in Figure~\ref{fig:main_framework}. We first formalize the problem through a Bayesian lens (Section~\ref{sec:method:bayesian}), deriving an objective that maximizes the mutual information between actions and instructions.
We then present our architecture, which uses \textbf{Latent Action Queries} to instantiate this decomposition (Section~\ref{sec:method:architecture}), and detail our dual-branch training strategy (Section~\ref{sec:method:training}).

\subsection{Objective Formulation}\label{sec:method:bayesian}

As established in Section~\ref{subsec:information_collapse}, standard VLA training on goal-driven datasets leads to information collapse where $\pi(a \mid v, \ell) \to p(a \mid v)$. To counteract this, we propose to regularize the policy by maximizing the conditional Pointwise Mutual Information (PMI) between the action and the instruction.
This objective can be formulated as maximizing the Log-Likelihood Ratio (LLR) between the posterior policy and the vision-only prior:
\begin{equation}
    \mathcal{L}_{\text{LLR}} = \log \frac{\pi(a \mid v, \ell)}{p(a \mid v)} = \log p(\ell \mid a, v) - \log p(\ell \mid v).
\end{equation}
The detailed derivation is provided in Appendix~\ref{app:derivation}.
This formulation requires us to simultaneously model the posterior $\pi(a \mid v, \ell)$ and the prior $p(a \mid v)$. In the following sections, we describe how \textbf{\methodname} efficiently instantiates these two distributions using a shared architecture with Latent Action Queries.

\subsection{Latent Action Queries}\label{sec:method:architecture}

To efficiently instantiate the proposed Bayesian decomposition within a unified VLM architecture, we introduce \textbf{Latent Action Queries}.
We extend the VLM vocabulary with $K=64$ learnable tokens, denoted as $\mathcal{Q} = \{ \texttt{<|action\_1|>}, \dots, \texttt{<|action\_K|>} \}$.
Correspondingly, the VLM's embedding layer is expanded with $K$ learnable embedding vectors. These vectors are learned to aggregate task-relevant information from the preceding vision and language tokens into a compact latent representation for action execution.

These queries function as a dedicated bottleneck at the interface between the VLM (e.g., Qwen3-VL \citep{Qwen3-VL_2025_arXiv}) and the continuous action head (a Diffusion Transformer \citep{DiT_2023_ICCV}). 
Crucially, while the VLM typically processes the full sequence, the bottleneck is enforced by exclusively forwarding the hidden states of these query tokens, $\mathbf{H}_{\mathcal{Q}} \in \mathbb{R}^{K \times D}$, to the action expert.
This contrasts with recent VLA architectures such as $\pi_0$~\citep{PI0_2024_arXiv} and GR00T \citep{GR00T_2025_arXiv,GR00T_N1.6}, which typically feed the hidden states of all input tokens to the action head.

This design choice is critical: by leveraging the causal masking inherent in decoder-only VLMs, we can precisely control the information encoded in $\mathbf{H}_{\mathcal{Q}}$ simply by changing the position of $\mathcal{Q}$ in the input sequence. This flexibility enables the strict separation of vision-only and vision-language contexts required for our dual-branch strategy.

\subsection{Dual-Branch Training Framework}\label{sec:method:training}

We propose a training paradigm with two parallel branches sharing the same VLM weights but different input.

\noindent\textbf{1. Priori Branch (Vision-Only).}
To estimate the prior $p(a \mid v)$, we construct the input sequence as:
\begin{equation}
    \text{Input}_{\text{prior}} = [v, \mathcal{Q}, \ell].
\end{equation}
Due to the causal attention mask of the decoder-only VLM, the tokens in $\mathcal{Q}$ can attend to the visual observation $v$ but \textit{cannot} attend to the language instruction $\ell$ (which appears later). Thus, the hidden states $\mathbf{H}_{\mathcal{Q}}^{\text{prior}}$ encode purely visual information.
We detach $\mathbf{H}_{\mathcal{Q}}^{\text{prior}}$ from the computation graph when optimizing $\mathcal{L}_{\text{prior}}$ to ensure that the gradient updates for the vision-only prior are confined to the DiT action head, preventing the shared VLM backbone from learning visual shortcuts.
We use these features to predict the action $a$, optimizing a flow-matching loss $\mathcal{L}_{\text{prior}}$ to learn the dataset's inherent action bias.

\noindent\textbf{2. Posteriori Branch (Vision + Language).}
To estimate the true policy $\pi(a \mid v, \ell)$, we arrange the input as:
\begin{equation}
    \text{Input}_{\text{post}} = [v, \ell, \mathcal{Q}].
\end{equation}
Here, $\mathcal{Q}$ appears after $\ell$, allowing it to attend to both vision and language. The resulting hidden states $\mathbf{H}_{\mathcal{Q}}^{\text{post}}$ encode the full context. We optimize a main flow-matching loss $\mathcal{L}_{\text{main}}$ to learn the expert action.

\input{tab/simpler}

\noindent\textbf{3. Maximizing the Likelihood Ratio.}
In addition to action prediction, we explicitly optimize the LLR objective. We treat the VLM's language modeling loss as a proxy for $\log p(\ell \mid \dots)$.
Specifically, in the Priori Branch, the language tokens $\ell$ attend to $[v, \mathcal{Q}]$. Since $\mathcal{Q}$ encodes the action information $a$ (via the prior), the probability of generating $\ell$ in this branch approximates $p(\ell \mid v, a_{\text{prior}})$.
In the Posteriori Branch, we can compute a baseline $p(\ell \mid v)$ by detaching gradients.
However, a more direct and numerically stable approach is to maximize the difference in log-probabilities of the language tokens between the two branches. We define the LLR loss as:
\begin{equation}
    \label{eq:llr}
    \mathcal{L}_{\text{LLR}} = \log p(\ell \mid v, \mathbf{H}_{\mathcal{Q}}^{\text{prior}}) - \text{sg}\left( \log p(\ell \mid v) \right),
\end{equation}
where $\text{sg}(\cdot)$ denotes the stop-gradient operator. We maximize this term (minimize $-\mathcal{L}_{\text{LLR}}$) to force the action representations $\mathbf{H}_{\mathcal{Q}}$ to carry information that explains $\ell$. 
The stop-gradient operation is employed to prevent the model from trivially maximizing the ratio by degrading the baseline $p(\ell \mid v)$ (i.e., damaging the VLM's general language capabilities) rather than improving the numerator.

\subsection{Total Training Objective}

We train the action decoder using the Rectified Flow Matching objective~\citep{liu2022flow, GR00T_2025_arXiv}. Specifically, we apply this objective to both the Priori Branch (conditioned on $\mathbf{H}_{\mathcal{Q}}^{\text{prior}}$) and the Posteriori Branch (conditioned on $\mathbf{H}_{\mathcal{Q}}^{\text{post}}$).
Given a condition $\mathbf{C} \in \{\mathbf{H}_{\mathcal{Q}}^{\text{post}}, \mathbf{H}_{\mathcal{Q}}^{\text{prior}}\}$, the flow-matching loss is defined as:
\begin{equation}
\label{eq:flow_matching}
    \mathcal{L}_{\text{FM}}(\psi; \mathbf{C}) = \mathbb{E}_{t, \mathbf{a}_0, \mathbf{a}_1} \left[ || v_\psi(\mathbf{a}_t, t, \mathbf{C}) - (\mathbf{a}_1 - \mathbf{a}_0) ||^2 \right],
\end{equation}
where $v_\psi$ is the Diffusion Transformer (DiT) predicting the velocity field, $\mathbf{a}_1$ is the ground truth action trajectory, $\mathbf{a}_0 \sim \mathcal{N}(0, I)$ is sampled from a standard Gaussian, and $\mathbf{a}_t = (1-t)\mathbf{a}_0 + t\mathbf{a}_1$ represents the interpolated state at timestep $t \in [0, 1]$.

The final training loss combines the action prediction losses from both branches with the LLR maximization term:
\begin{equation}
    \label{eq:total_loss}
    \mathcal{L}_{\text{total}} = (1 - \lambda) \mathcal{L}_{\text{FM}}(\psi; \mathbf{H}_{\mathcal{Q}}^{\text{post}}) + \lambda \mathcal{L}_{\text{FM}}(\psi; \mathbf{H}_{\mathcal{Q}}^{\text{prior}}) - \beta \mathcal{L}_{\text{LLR}},
\end{equation}
where $\lambda$ balances the contribution of the prior and posterior action losses, and $\beta$ controls the strength of the LLR regularization.
We set $\lambda=0.3$ and $\beta=0.1$ in our experiments.
During inference, we exclusively execute the Posteriori Branch to obtain $\mathbf{H}_{\mathcal{Q}}^{\text{post}}$ and generate actions via the DiT. 
This ensures that our method incurs no additional computational overhead compared to standard VLA baselines at test time.

%% file: tab/simpler.tex
\definecolor{navyblue}{HTML}{0071BC}

\begin{table*}[ht]
  \renewcommand{\arraystretch}{1.2} 
  \centering
  \caption{
    \textbf{Results of evaluating the VLA models with the WidowX robot in the SimplerEnv simulation environment}. We highlight the best results in \textbf{bold} and the second-best results with \underline{underline}.
    }
  \begin{adjustbox}{width=\linewidth}
  \rowcolors{24}{white}{gray!15}
  \begin{tabular}{l c c c c c}
    \toprule
    \textbf{Method}
     & \makecell[c]{\textbf{Put Spoon} \\ \textbf{on Towel}} 
     & \makecell[c]{\textbf{Put Carrot} \\ \textbf{on Plate}} 
     & \makecell[c]{\textbf{Stack Green Block} \\ \textbf{on Yellow Block}} 
     & \makecell[c]{\textbf{Put Eggplant} \\ \textbf{in Yellow Basket}} 
     & \textbf{Average} \\
    \midrule
    OpenVLA-OFT~\citep{OpenVLA-OFT_2025_arXiv}     & 34.2  & 30.0  & 30.0  & 72.5  & 41.8 \\
    RoboVLM~\citep{RoboVLM_2024_arXiv}         & 50.0  & 37.5  & 0.0   & 83.3  & 42.7 \\ 
    Magma~\citep{yang2025magma}                & 37.5  & 29.2  & 20.8  & 91.7  & 44.8 \\  
    CogACT~\citep{CogACT_2024_arXiv}           & 71.7 &  50.8  & 15.0 & 67.5 & 51.3 \\
    SpatialVLA~\citep{Spatialvla_2025_arXiv}      & 20.8  & 20.8  & 25.0  & 70.8  & 34.4 \\
    TraceVLA~\citep{TraceVLA_2025_arXiv}        & 12.5  & 16.6  & 16.6  & 65.0  & 27.7 \\
    VideoVLA~\citep{VideoVLA_2025_NeurIPS}        & 75.0 & 20.8   & 45.8 & 70.8 & 53.1 \\

    \midrule
    $\pi_0$~\citep{PI0_2024_arXiv}         & 29.2 & 62.5 & 29.2 & 91.6 & 53.1 \\
    $\pi_{0.5}$~\citep{PI05_2025_arXiv} & 49.3 & 64.7 & 44.7 & 69.7 & \underline{57.1} \\
    Isaac-GR00T-N1.6-Bridge~\citep{GR00T_N1.6}   & 64.5 & 65.5 & 5.5 & 93.0 & \underline{57.1} \\
    \midrule
    QwenGR00T (Baseline) + Qwen3-VL-4B~\citep{starvla_2025}  & 87.5 &  50.0 & 29.2 & 54.2 & 55.2 \\
    \rowcolor{gray!30} \textbf{\methodname} + Qwen3-VL-4B  &  89.6 &  63.8 & 33.3 & 79.2 & \textbf{66.5} \\
    
    \bottomrule
  \end{tabular}
  \end{adjustbox}
  \vspace{-0.5em}
  \label{tab:simplerenv}
\end{table*}

%% file: sec/4_experiment.tex
\section{Experiment}\label{sec:experiment}

To comprehensively evaluate the effectiveness of \methodname, we conduct extensive experiments on two main simulation benchmarks, SimplerEnv and RoboCasa, with additional LIBERO results reported in Appendix~\ref{app:libero}. We also evaluate \methodname in two real-world robot settings: colored-block pick-and-place and vegetable pick-and-place.
Our training pipeline is built upon the StarVLA framework~\citep{starvla_2025}, distributed across 8 NVIDIA H100 GPUs, and strictly follows its default training protocols to ensure fair comparison. 
In our experiments, \methodname is instantiated on the QwenGR00T architecture from StarVLA.
We employ the AdamW optimizer~\citep{AdamW_2017_ICLR} initialized with a learning rate of 1e-5 and a cosine annealing schedule. 
System-level optimizations include DeepSpeed ZeRO-2 \citep{rasley2020deepspeed}, gradient clipping at a norm of 1.0, and no gradient accumulation.
All baseline performance metrics are obtained from their original papers or other peer-reviewed publications. To ensure a fair comparison, the training datasets for these baselines encompass the data used in our experiments.

\subsection{Experiments on SimplerEnv}\label{subsec:simpler_exp}
We utilize two large-scale subsets from the Open X-Embodiment (OXE) dataset: BridgeDataV2 \citep{Bridgedatav2_2023_CoRL} and Fractal \citep{RT-1_2022_arXiv}. 
The model is fine-tuned for 50k steps on a cluster of 8 GPUs (batch size 16 per device). 
This benchmark includes four manipulation tasks: ``Put spoon on towel'', ``Put carrot on plate'', ``Stack green cube on yellow cube'', and ``Put eggplant in yellow basket''. 
For each task, we evaluate the VLA policies using the official evaluation scripts provided by the SimplerEnv repository~\citep{SimplerEnv_2024_CoRL}. 
To mitigate the effects of randomness, we run 480 independent trials and report the average performance (Avg@480).

\input{tab/robocasa_avg}

The results are summarized in Table~\ref{tab:simplerenv}. 
\methodname consistently outperforms comparison baselines, achieving a state-of-the-art average success rate of 66.5\%. 
Notably, compared to the direct baseline QwenGR00T (55.2\%) built on the same StarVLA framework, our method delivers an absolute improvement of 11.3\%, validating that the performance gain stems from our proposed Bayesian decomposition rather than the base architecture.
Significant improvements are observed in tasks requiring precise object identification and manipulation, such as \textit{``Put Carrot on Plate''} (+13.6\%) and \textit{``Put Eggplant in Yellow Basket''} (+15.0\%).
Furthermore, \methodname surpasses other recent strong competitors, including the flow-matching-based $\pi_{0.5}$ (57.1\%) and the dual-system Isaac-GR00T-N1.6 (57.1\%).
These results confirm that by explicitly optimizing the mutual information between language and action, \methodname effectively mitigates the vision shortcut. 
Fundamentally, this validates that our approach prevents the policy from collapsing into a spurious vision-only prior $p(a|v)$ caused by dataset determinism, and instead compels the model to learn the true causal dependency of actions on language instructions.


\subsection{Experiments on RoboCasa}

We evaluate our method on the RoboCasa GR1 Tabletop Manipulation Benchmark \citep{RoboCasa_2024_RSS}, which consists of 24 diverse manipulation tasks.
These tasks feature complex interactions with articulated objects and varied geometries, exemplified by specific tasks like ``PnPBottleToCabinetClose'' and ``PnPCanToDrawerClose'', as well as scenarios involving appliances like microwaves and toasters.
For training, we utilize the Humanoid Robot Tabletop Manipulation subset from the PhysicalAI-Robotics-GR00T-X-Embodiment-Sim \citep{GR00T_2025_arXiv} dataset.
All other settings follow Section~\ref{subsec:simpler_exp}.
To guarantee statistical significance, we evaluate each task using 50 independent trials and report the average success rate (Avg@50).

The quantitative results on RoboCasa are presented in Table~\ref{tab:robocasa_main_tab}.
\methodname achieves a state-of-the-art average success rate of 52.6\%, surpassing all competing baselines including QwenOFT (48.8\%), Isaac-GR00T N1.5 (48.2\%), and the direct baseline QwenGR00T (47.8\%).
Notably, our method substantially outperforms the VisionOnly baseline (44.7\%), demonstrating the effectiveness of language grounding.
Detailed per-task results can be found in Appendix~\ref{app:robocasa}.



\subsection{Experiments on Real World}
We evaluate \methodname on a real-world robotic setup using a Franka Research 3 arm. We design a ``\textbf{Pick and Place All Blocks}'' task involving colored blocks to assess instruction following and OOD generalization. 
We intentionally focus on pick-and-place settings because they provide a transparent testbed for separating instruction following from dexterous control: when there is only one plausible target in the scene, most VLA policies can succeed by relying primarily on manipulation competence, whereas selecting the correct target color or object category reveals whether the policy truly uses the language instruction.
Notably, while the training demonstrations only contain single-block manipulation, we evaluate the model in scenarios requiring it to sequentially pick up multiple blocks (1, 2, or 3) and place them into the box.
This setting serves as a rigorous test for the model's generalization capabilities and long-horizon planning skills.
Detailed experimental setup is provided in Appendix~\ref{app:real_world}.

As shown in Table~\ref{tab:real_robot}, our model achieves a success rate of 25/30 on in-domain colors (single block), outperforming the QwenGR00T baseline (21/30).
Consistent with our SimplerEnv findings (where tasks like ``Stack Green Cube'' requiring precise low-level control showed smaller gains), the improvement in the in-domain setting is moderate. 
This is expected, as simple block manipulation primarily challenges the robot's precise motor control rather than its high-level language understanding.
However, on the challenging out-of-distribution (OOD) task involving the unseen Red block, our model achieves a success rate of 9/30, significantly surpassing QwenGR00T, which struggles with a success rate of only 2/30.
This contrast highlights that while \methodname maintains competitive low-level control capabilities, its primary advantage lies in enhanced instruction following and generalization to novel semantic concepts.

\input{tab/realworld}

To further evaluate the method on objects with more diverse shapes and appearances, we additionally conduct a real-world ``\textbf{Vegetable Pick and Place}'' task using the same robot setup as the ``Pick and Place All Blocks'' task. We collect 100 demonstrations for each of four vegetables (eggplant, pepper, carrot, and cucumber) in the same scene, train \methodname and QwenGR00T under the same protocol, and evaluate each method over 30 trials per vegetable.
As shown in Table~\ref{tab:real_robot_vegetable}, \methodname achieves an overall success rate of 80.8\% (97/120), substantially outperforming QwenGR00T's 59.2\% (71/120). The consistent gains across all four objects indicate that the benefit of language-grounded action learning transfers beyond colored-block manipulation to real objects with varied geometry and appearance.
Nevertheless, these real-world experiments remain limited to relatively simple pick-and-place behaviors and do not fully validate contact-rich or dexterous manipulation, which we leave for future work.

\input{tab/realworld_vegetable}

\begin{figure*}[!h]
    \centering
    \includegraphics[width=1\textwidth]{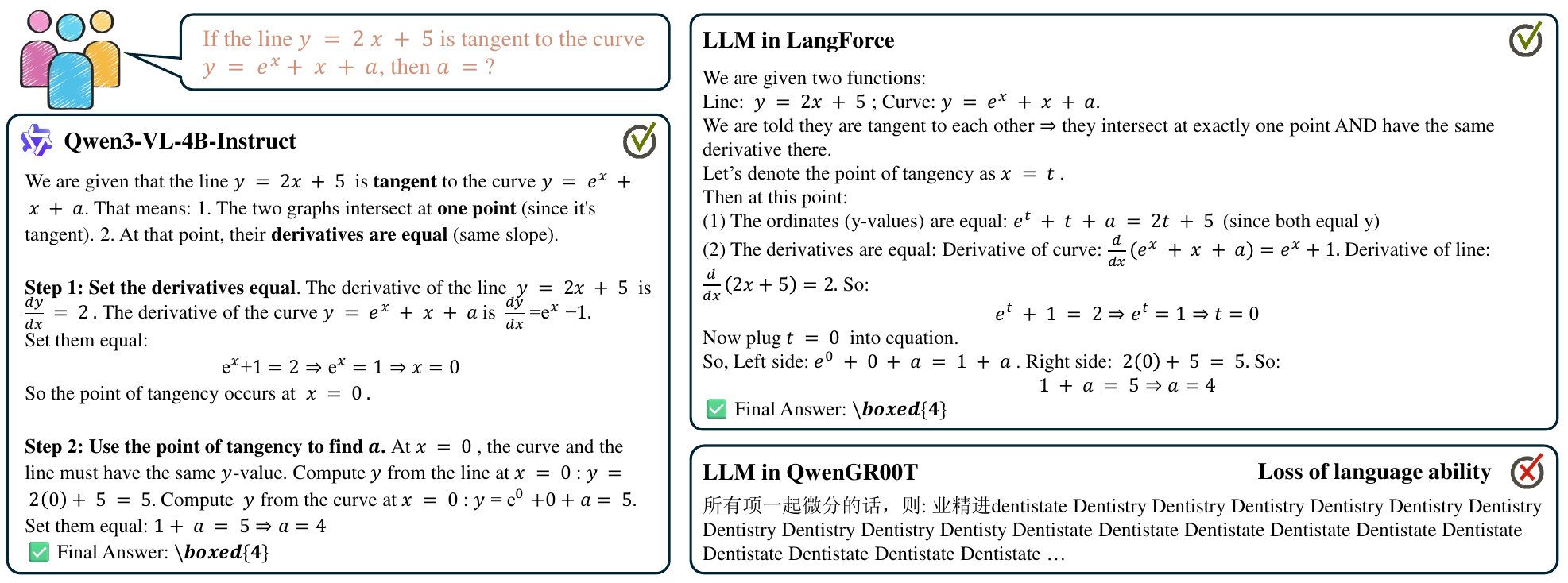}
    \caption{\textbf{Qualitative comparison of general multimodal reasoning.} We present a case where the model is asked to solve a mathematical problem. The standard VLA baseline (QwenGR00T) suffers from catastrophic forgetting; while the text before the comma implies ``differentiating all terms together'', the subsequent output degenerates into repetitive and meaningless gibberish (bottom right). In contrast, \methodname (top right) retains the VLM's original reasoning and language generation capabilities (left), successfully solving the problem.}
    \label{fig:qa}
\end{figure*}

\subsection{Preservation of General Capabilities}

A prevalent concern in the field is that fine-tuning VLMs for robotic action generation (VLA training) often compromises the model's foundational reasoning and multimodal understanding, leading to a degradation of general conversational abilities~\citep{ChatVLA2_2025_arXiv,xu2025seeing,VLM2VLA_2025_arXiv,TwinBrainVLA_2026_arXiv}.
ChatVLA~\citep{ChatVLA2_2025_arXiv} attributes this to \textit{spurious forgetting}, where robot training overwrites critical visual-text alignments, and \textit{task interference} between control and understanding objectives.

We observe similar degradation in our baseline: as illustrated in Figure~\ref{fig:qa} and Figure~\ref{fig:qa2}, the standard QwenGR00T model loses the ability to converse coherently, even when prompted with pure text inputs.
In stark contrast, \methodname remarkably preserves these capabilities when queried with language instructions.
This is because our LLR objective enforces a strong dependency on language for action generation, preventing the linguistic representations from collapsing or becoming redundant.
We provide a detailed discussion of this phenomenon and mechanisms in Appendix~\ref{app:general_capabilities}.

\subsection{Ablation Studies}

We conduct ablation studies on SimplerEnv to validate the contributions of individual components in \methodname. All experiments utilize the Qwen3-VL-4B backbone, and results are presented in Table~\ref{tab:ablation}.

\textbf{Effectiveness of Bayesian Decomposition.} 
Comparing the full \methodname (63.5\%) with the ``+ Action Query'' ablation (57.5\%), we observe a significant performance boost (+6.0\%). 
This indicates that while the architectural changes provide some benefit, the core improvement stems from our dual-branch Bayesian learning objective. 
By explicitly modeling and maximizing the pointwise mutual information (PMI) between instructions and actions, the model effectively overcomes the vision shortcut, validating the central hypothesis of this work.
Additionally, we provide ablation studies on the hyperparameters $\lambda$ and $\beta$ in Appendix~\ref{app:ablation_lambda_beta}.

\input{tab/ablation}

\textbf{Potential of Latent Action Queries.}
Even without the dual-branch definition, introducing Latent Action Queries (``+ Action Query'') improves upon the QwenGR00T baseline (55.2\% $\to$ 57.5\%). 
This suggests that Latent Action Queries function as a promising architectural inductive bias. 
Unlike standard approaches that feed full sequences of vision and language token embeddings into the action decoder, our query-based mechanism forces the VLM to compress and summarize task-relevant information into a compact set of latent tokens.
From a computational perspective, this design is highly efficient. 
It decouples the complexity of the Diffusion Transformer (DiT) from the length of the VLM input context. 
Specifically, the complexity of condition processing in the DiT is reduced from $O(N^2)$ (scaling with the massive number of vision-language tokens $N$) to $O(K^2)$ (scaling with the small, constant number of query tokens $K$), thereby streamlining the action generation process.

We also investigate the impact of the number of latent action queries on model performance. Detailed analysis and results are provided in Appendix~\ref{app:ablation_query}.

%% file: tab/robocasa_avg.tex
\begin{table}[!t]
    \centering
    \small
    \renewcommand{\arraystretch}{1.3} 

    \caption{
      \textbf{Results of evaluating the VLA models with the GR1 robot in the RoboCasa Tabletop simulation environment}. The results for Isaac-GR00T N1.5 and Isaac-GR00T N1.6 are sourced from the official Isaac-GR00T github repository~\citep{GR00T_2025_arXiv}. The results for the first four baseline methods are sourced from the official starVLA experiments~\citep{starvla_2025}. Performance on all 24 tasks can be found in Table~\ref{tab:robocasa_supp_tab} of the Appendix.
    }
    \begin{adjustbox}{width=0.96\columnwidth}
    \begin{tabular}{l c c c c}
        \toprule
        \rowcolor{white} 
        {Method} & 
        {\scriptsize \makecell{QwenFAST\\+Qwen3VL}} &
        {\scriptsize \makecell{QwenGR00T\\+Qwen3VL}} & 
        {\scriptsize \makecell{QwenPI\\+Qwen3VL}} & 
        {\scriptsize \makecell{QwenOFT\\+Qwen3VL}} \\
        \midrule
        \textbf{Average} & 39.0 & 47.8 & 43.9 & \underline{48.8} \\
        \midrule
        \rowcolor{white} 
        {Method} & 
        {\scriptsize \makecell{Isaac-GR00T\\N1.5}} & 
        {\scriptsize \makecell{Isaac-GR00T\\N1.6}} & 
        {\scriptsize \makecell{VisionOnly \\QwenGR00T}} & 
        {\scriptsize \makecell{\methodname\\+Qwen3VL}} \\
        \midrule
        \textbf{Average} & 48.2 & 47.6 & 44.7 & \textbf{52.6} \\
        \bottomrule
    \end{tabular}
    \end{adjustbox}
    \label{tab:robocasa_main_tab}
    \vspace{-1em}
\end{table}

%% file: tab/realworld.tex
\begin{table}[t]
    \centering
    \scriptsize
    \caption{\textbf{Real World Pick and Place All Blocks experiment.} Success rates (\%) of the pick-and-place task on the Franka Research 3 robot. The headers 1, 2, and 3 denote the number of blocks in the scene that need to be placed into the box. We evaluate performance for both \textbf{In-Domain} and \textbf{Out-of-Domain} (unseen object colors).}
    \label{tab:real_robot}
    \begin{adjustbox}{width=1\columnwidth}
    \begin{tabular}{lccccc}
        \toprule
        \multirow{2}{*}{\textbf{Method}} & \multicolumn{3}{c}{\textbf{In-Domain}} & \multicolumn{2}{c}{\textbf{Out-of-Domain}}\\
        \cmidrule(lr){2-4} \cmidrule(lr){5-6}
        & 1 & 2 & 3 & 1 & 2 \\
        \midrule
        OpenVLA~\citep{OpenVLA-OFT_2025_arXiv} & 6/30 & 2/30 & 0/30 & 0/30 & 0/30\\
        $\pi_{0.5}$~\citep{PI05_2025_arXiv} & \textbf{25/30} & \textbf{19/30} & \underline{13/30} & \underline{7/30} & \underline{1/30}\\
        QwenGR00T & \underline{21/30} & 14/30 & 9/30 & 2/30 & 0/30 \\
        \textbf{\methodname{} (Ours)} & \textbf{25/30} & \underline{18/30} & \textbf{14/30} & \textbf{9/30} & \textbf{3/30}\\
        \bottomrule
    \end{tabular}
    \end{adjustbox}
    \vspace{-1em}
\end{table}

%% file: tab/realworld_vegetable.tex
\begin{table}[t]
    \centering
    \scriptsize
    \caption{\textbf{Real-world vegetable Pick and Place experiment.} We use the same Franka Research 3 setup as the ``Pick and Place All Blocks'' task and evaluate each method over 30 trials per vegetable.}
    \label{tab:real_robot_vegetable}
    \begin{adjustbox}{width=1\columnwidth}
    \setlength{\tabcolsep}{13pt} 
    \begin{tabular}{lcc}
        \toprule
        \textbf{Vegetable} & \textbf{\methodname{} (Ours)} & \textbf{QwenGR00T} \\
        \midrule
        Eggplant & 24/30 (80.0\%) & 16/30 (53.3\%) \\
        Pepper & 25/30 (83.3\%) & 21/30 (70.0\%) \\
        Carrot & 27/30 (90.0\%) & 18/30 (60.0\%) \\
        Cucumber & 21/30 (70.0\%) & 16/30 (53.3\%) \\
        \midrule
        \textbf{Overall} & \textbf{97/120 (80.8\%)} & 71/120 (59.2\%) \\
        \bottomrule
    \end{tabular}
    \end{adjustbox}
    \vspace{-1em}
\end{table}

%% file: tab/ablation.tex
\definecolor{navyblue}{HTML}{0071BC}

\begin{table}[!t]
  \centering
  \caption{
    \textbf{Ablation study on SimplerEnv}. All experiments are based on the Qwen3-VL-4B backbone. We compare the baseline QwenGR00T, the addition of Latent Action Queries, and the full \methodname{} to validate the contributions of each component.
  }
  \begin{adjustbox}{width=1\columnwidth}
  
  \begin{tabular}{l c c c c c}
    \toprule
    \textbf{Method}
     & \makecell[c]{\textbf{Put Spoon} \\ \textbf{on Towel}} 
     & \makecell[c]{\textbf{Put Carrot} \\ \textbf{on Plate}} 
     & \makecell[c]{\textbf{Stack} \\ \textbf{Block}} 
     & \makecell[c]{\textbf{Eggplant} \\ \textbf{in Basket}} 
     & \textbf{Avg} \\
    \midrule
    QwenGR00T  & 87.5 &  50.0 & 29.2 & 54.2 & 55.2 \\
    + Action Query  & 74.6 &  58.3 & 29.2 & 67.9 & \underline{57.5} \\
    \textbf{\methodname}  &  89.6 &  63.8 & 33.3 & 79.2 & \textbf{66.5} \\
    
    \bottomrule
  \end{tabular}
  \end{adjustbox}
  \vspace{-1em}
  \label{tab:ablation}
\end{table}

%% file: sec/6_related.tex
\section{Related Work}\label{sec:related}

We build our work upon the following rigorous foundations:

\noindent\textbf{Vision-Language-Action Dataset and Benchmark.}

The advancement of generalist robot policies relies heavily on large-scale datasets and rigorous benchmarks.
LIBERO~\citep{LIBERO_2023_NeurIPS} pioneered the systematic study of knowledge transfer in lifelong robot learning.
To scale up real-world data, BridgeData V2~\citep{Bridgedatav2_2023_CoRL} provided diverse interaction trajectories on low-cost hardware.
This effort was expanded by Open X-Embodiment (OXE)~\citep{OXE_2024_ICRA}, which aggregated data across 22 robot embodiments, and Droid~\citep{Droid_2024_arXiv}, which further increased diversity with distributed data collection.
For scalable evaluation, RoboCasa~\citep{RoboCasa_2024_RSS} introduced a large-scale simulation framework with realistic kitchen environments, while SimplerEnv~\citep{SimplerEnv_2024_CoRL} provided a simulated evaluation proxy to correlate with real-world performance, addressing the reproducibility crisis in physical evaluation.
More recently, RoboTwin 2.0~\citep{RoboTwin2_2025_arXiv} offered a unified benchmark for bimanual manipulation with automated data generation, and AgiBot-World~\citep{Agibot_2025_arXiv} scaled training data to over 1 million trajectories with human-in-the-loop verification.

\noindent\textbf{Vision-Language-Action Models.} 



To bridge the gap between semantic understanding and physical control, Vision-Language-Action (VLA) models have emerged as a dominant paradigm.
Early works like OpenVLA~\citep{OpenVLA_2024_CoRL} and its variant OpenVLA-OFT~\citep{OpenVLA-OFT_2025_arXiv} fine-tune large language models for robotic control.
Further architectural innovations include using diffusion transformers~\citep{CogACT_2024_arXiv, RDT-1B_2025_ICLR} or dual-system designs like the GR00T series~\citep{GR00T_2025_arXiv,GR00T_N1.5_2025,GR00T_N1.6}, which couples a VLM with a diffusion head.
Recently, the $\pi_0$ series~\citep{PI0_2024_arXiv, PI05_2025_arXiv, FAST_2025_arXiv} utilizes data from multiple robots, high-level semantic prediction, web data, and other sources to enable broadly generalizable real-world robotic manipulation

Other approaches like X-VLA~\citep{X-VLA_2025_arXiv} introduce embodiment-specific soft prompts to facilitate cross-embodiment generalization. By learning separate sets of embeddings for each data source, X-VLA effectively leverages heterogeneous robot data with minimal additional parameters.
SpatialVLA~\citep{Spatialvla_2025_arXiv} argues that spatial understanding is central to manipulation, introducing Ego3D Position Encoding and Adaptive Action Grids to inject 3D information and learn transferable spatial action knowledge.
3D-Mix~\citep{3dmix_2026_arXiv} further studies VGGT-based 3D feature integration and proposes a plug-and-play gated fusion module that adaptively balances 2D semantic and 3D geometric features for VLA models.
Then, VideoVLA~\citep{VideoVLA_2025_NeurIPS} explores transforming video generation models into robot manipulators.
By jointly predicting action sequences and future visual outcomes, it leverages the ``visual imagination'' of generative models to enhance generalization across novel tasks and objects.
TwinBrainVLA~\citep{TwinBrainVLA_2026_arXiv} mitigates catastrophic forgetting by coordinating a frozen generalist ``Left Brain'' for semantic understanding and a trainable specialist ``Right Brain'' for sensorimotor learning, effectively balancing high-level reasoning with low-level control. 
BayesVLA~\citep{xu2025seeing} also employs Bayesian decomposition to improve instruction following. It utilizes a two-stage framework: first training a vision-conditioned prior on large-scale vision-action pairs, and then freezing this prior to train a language-conditioned posterior. However, this design strictly requires a separated training process where the vision component must be frozen in the second stage. In contrast, \methodname enables single-stage, end-to-end training.

Overall, other VLA approaches lack systematic solutions to the vision shortcut problem, where models ignore language instructions in goal-driven datasets.

%% file: sec/7_conclusion.tex
\section{Discussion}

Our analysis in previous sections suggests that the deterministic mapping between visual observations and language instructions in goal-driven datasets may facilitate the vision shortcut. Consequently, we advocate for:

\infobox{
Prioritizing \textbf{ambiguous scenarios} during data collection to naturally compel models to rely on language for disambiguation.
}

This data-centric direction is a long-term and fundamental solution. At the same time, a large amount of existing VLA data has already been collected under approximately goal-deterministic settings, and recollecting or re-annotating such data at scale is expensive. \methodname provides a way to make better use of these existing datasets. Moreover, even with broader data, shortcut behavior can still arise whenever optimization exploits local scene-to-task correlations. An explicit objective that contrasts the language-conditioned posterior with a vision-only prior therefore remains valuable for encouraging genuine language use.

Additionally, for further insights on data collection, leveraging human data, and connections to World Models, we refer readers to the detailed discussion in Appendix~\ref{sec:discussion}.

\section{Limitation}
\label{sec:limitation}

While \methodname offers significant improvements in robustness, the dual-branch architecture introduces a limitation regarding computational overhead during training. 
Since the model must compute both the Priori and Posteriori branches, the computational cost per iteration theoretically increases. 
However, we note that the visual input prefix is identical for both branches, and the number of visual tokens vastly outnumbers that of the language and latent action query tokens. 
By employing a prefix prefill strategy to compute and reuse the visual representations (e.g., vision encoder outputs) for both branches, the actual increase in training time is marginal. 
Thus, the additional computational overhead remains within a completely acceptable range.

Our real-world experiments are limited to relatively simple pick-and-place settings that primarily evaluate instruction following rather than dexterous manipulation. This aligns with the goal of improving language grounding rather than low-level manipulation skill. Using simple tasks makes failures easier to attribute to insufficient instruction following instead of execution errors.

\section{Conclusion}\label{sec:conclusion}

In this work, we identify the \textbf{vision shortcut} in VLA training, where policies rely on visual correlations rather than language under approximately goal-deterministic data.
We analyze this failure mode through a Bayesian and information-theoretic lens and introduce \textbf{\methodname}, a post-training framework that contrasts a language-conditioned posterior with a vision-only prior to encourage instruction-specific action information.
With Latent Action Queries and a shared dual-branch architecture, \methodname improves language grounding without inference overhead.
Across SimplerEnv, RoboCasa, LIBERO, and two real-world pick-and-place settings, \methodname improves over VLA baselines in ambiguous and OOD scenarios.
These results highlight the importance of objectives that prevent shortcuts and preserve the causal role of language in action generation.
We hope this perspective encourages future VLA research to evaluate not only task success, but also whether policies genuinely condition actions on language.

%% file: sec/x_appendix.tex
\section{Derivation of the LLR Objective}\label{app:derivation}

In this section, we provide the derivation for the Log-Likelihood Ratio (LLR) objective used in \methodname.
Our core motivation is to maximize the learning signal discussed previously: the Conditional Pointwise Mutual Information (PMI) between the action $a$ and the language instruction $\ell$, given the visual observation $v$. 
The PMI is formally defined as:
\begin{equation}
    \text{PMI}(a, \ell \mid v) = \log \frac{\pi(a, \ell \mid v)}{p(a \mid v) p(\ell \mid v)}.
\end{equation}
By applying the chain rule of probability $\pi(a, \ell \mid v) = \pi(a \mid v, \ell) p(\ell \mid v)$, we can rewrite the PMI as the log-ratio between the posterior policy and the vision-only prior:
\begin{equation}
    \text{PMI}(a, \ell \mid v) = \log \frac{\pi(a \mid v, \ell) p(\ell \mid v)}{p(a \mid v) p(\ell \mid v)} = \log \frac{\pi(a \mid v, \ell)}{p(a \mid v)}.
\end{equation}
This formulation highlights that maximizing PMI is equivalent to maximizing the divergence between the language-conditioned policy and the vision-only prior, directly penalizing the collapse where $\pi(a \mid v, \ell) \approx p(a \mid v)$.

Alternatively, using the chain rule $\pi(a, \ell \mid v) = p(\ell \mid a, v) p(a \mid v)$, we arrive at the second form, which constitutes our practical LLR objective:
\begin{equation}
    \text{PMI}(a, \ell \mid v) = \log \frac{p(\ell \mid a, v) p(a \mid v)}{p(a \mid v) p(\ell \mid v)} = \log \frac{p(\ell \mid a, v)}{p(\ell \mid v)} = \log p(\ell \mid a, v) - \log p(\ell \mid v).
\end{equation}
This objective represents the difference between the log-likelihood of the instruction given the action and vision, and the log-likelihood of the instruction given vision alone. 
Maximizing this quantity compels the model to select actions $a$ that make the instruction $\ell$ significantly more probable than determining it from the visual context $v$ alone, thereby extracting the additional information required to solve ambiguous or out-of-distribution tasks.

\section{Additional Experiments}\label{app:additional_experiments}


\subsection{Additional Experiments on LIBERO}\label{app:libero}

\input{tab/libero}

We also evaluate \methodname{} on the LIBERO benchmark~\citep{LIBERO_2023_NeurIPS}.
Given that the training and testing environments in LIBERO are highly similar and current VLA research has largely saturated this benchmark (with baselines exceeding 95\%), our method yields comparable performance on the Spatial, Object, and Long suites.
However, on the \textbf{Goal} suite, \methodname{} achieves a success rate of 99.4\%, outperforming the Qwen3-VL-GR00T baseline (97.4\%) by \textbf{+2.0\%}.
As highlighted in Section~\ref{sec:motivation}, the Goal suite features significant visual ambiguity where multiple tasks share the same scene. This result empirically validates that our method effectively mitigates the vision shortcut, enabling the model to resolve ambiguity through robust instruction following.

To further substantiate this claim, we conducted a quantitative analysis of the conditional entropy $H(\ell \mid v)$ on the LIBERO Goal dataset. We approximated this metric by computing the Negative Log-Likelihood (NLL) of the ground-truth instructions given the visual observations across 40,000 samples.
The results are summarized in Table~\ref{tab:libero_nll}.

\begin{table}[h]
    \centering
    \small
    \caption{\textbf{Quantitative analysis of conditional entropy on LIBERO Goal.} We report the Negative Log-Likelihood (NLL) and Perplexity (PPL) of instructions given visual observations, serving as a proxy for $H(\ell \mid v)$. Higher NLL/PPL indicates that the model preserves the necessary uncertainty about the task given only vision, preventing the information collapse observed in baselines.}
    \label{tab:libero_nll}
    \begin{tabular}{l c c c}
        \toprule
        \textbf{Method} & \textbf{NLL (nats/token)} $\uparrow$ & \textbf{PPL ($\exp(\text{NLL})$)} $\uparrow$ & \textbf{Std Dev (NLL)} \\
        \midrule
        QwenGR00T (Baseline) & 8.51 & 4964.1 & 0.55 \\
        \textbf{\methodname{}} & \textbf{9.47} & \textbf{12964.9} & 0.54 \\
        \bottomrule
    \end{tabular}
\end{table}

\methodname achieves a significantly higher NLL (\textbf{9.47} nats/token) compared to the QwenGR00T baseline (\textbf{8.51} nats/token).
This difference is even more pronounced in terms of Perplexity (PPL), where \methodname reaches \textbf{12964.9} compared to the baseline's \textbf{4964.1}.
It is worth noting that the language structure in LIBERO Goal is highly repetitive (e.g., ``put the [object] in/on the [receptacle]''), which naturally encourages the model to fit these syntactic patterns.
Despite this, our method achieves a higher NLL and PPL, indicating that the uncertainty stems primarily from the task-specific nouns (objects and receptacles) rather than the sentence structure.
In the context of LIBERO Goal, where visual scenes are inherently ambiguous, the baseline's lower NLL implies it is ``overconfident'' in predicting these key nouns solely from vision. This confirms that standard training leads to a collapse in conditional entropy, where the model learns spurious correlations ($v \to \ell$) and ignores the actual need for language disambiguation.
In contrast, the higher NLL observed in \methodname indicates that our model prevents the pathological collapse of $H(\ell \mid v)$ observed in standard training, preserving uncertainty levels that align with the inherent ambiguity of visual scenes. By maintaining this necessary uncertainty, our method forces the policy to actively utilize the provided language instruction to resolve ambiguity, directly contributing to the superior performance on the Goal suite.

\subsection{Additional Experiments on RoboCasa}\label{app:robocasa}
More quantitative results on RoboCasa are presented in Table~\ref{tab:robocasa_supp_tab}.
Consistent with the empirical evidence presented in our motivation (Section~\ref{sec:motivation}), the VisionOnly baseline achieves a surprisingly high success rate of 44.7\%, lagging only slightly behind the standard QwenGR00T baseline (47.8\%). 
This observation reconfirms the prevalence of the vision shortcut in this benchmark, suggesting that a significant portion of tasks can be solved by relying solely on visual cues.
However, \methodname breaks this performance ceiling, achieving a state-of-the-art average success rate of 52.6\% and surpassing all competing baselines, including QwenOFT (48.8\%), Isaac-GR00T N1.5 (48.2\%), and Isaac-GR00T N1.6 (47.6\%).
Crucially, our method demonstrates substantial gains in tasks where the vision-only policy falters. 
For instance, in ``PnP Novel From Placemat To Bowl'', \methodname reaches 62.0\%, substantially surpassing both the VisionOnly baseline (32.0\%) and QwenGR00T (44.0\%).
These results indicate that maximizing the LLR objective successfully forces the policy to extract and utilize task-specifying information from language, rather than settling for local optima based on visual shortcuts.

\input{tab/app_realworld}

\subsection{Real-World Experiments Setup}\label{app:real_world}

\input{tab/robocasa}

\begin{figure*}[h]
    \centering
    \includegraphics[width=1\textwidth]{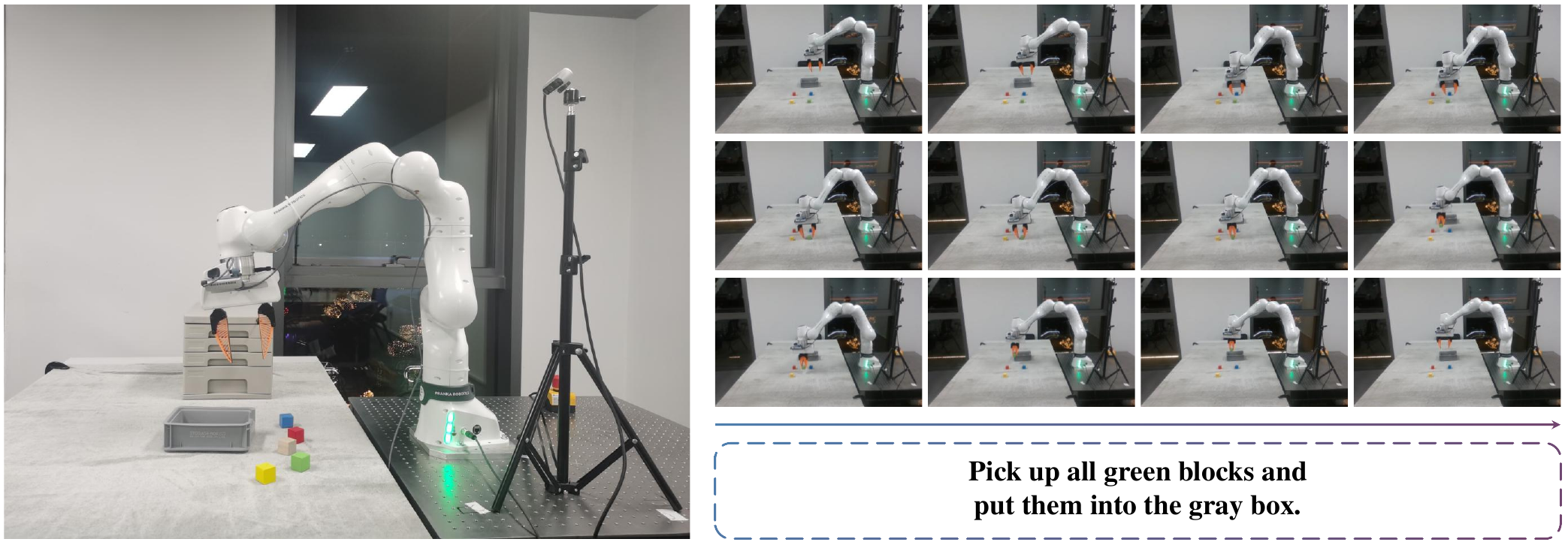}
    \caption{\textbf{Diagram of human expert data acquired through teleoperation.} We use Polymetis \citep{polymetis} for teleoperation and collect human expert data to support imitation learning for VLA
models.}
    \label{fig:real_world_show}
\end{figure*}

\begin{figure*}[h]
    \centering
    \includegraphics[width=1\textwidth]{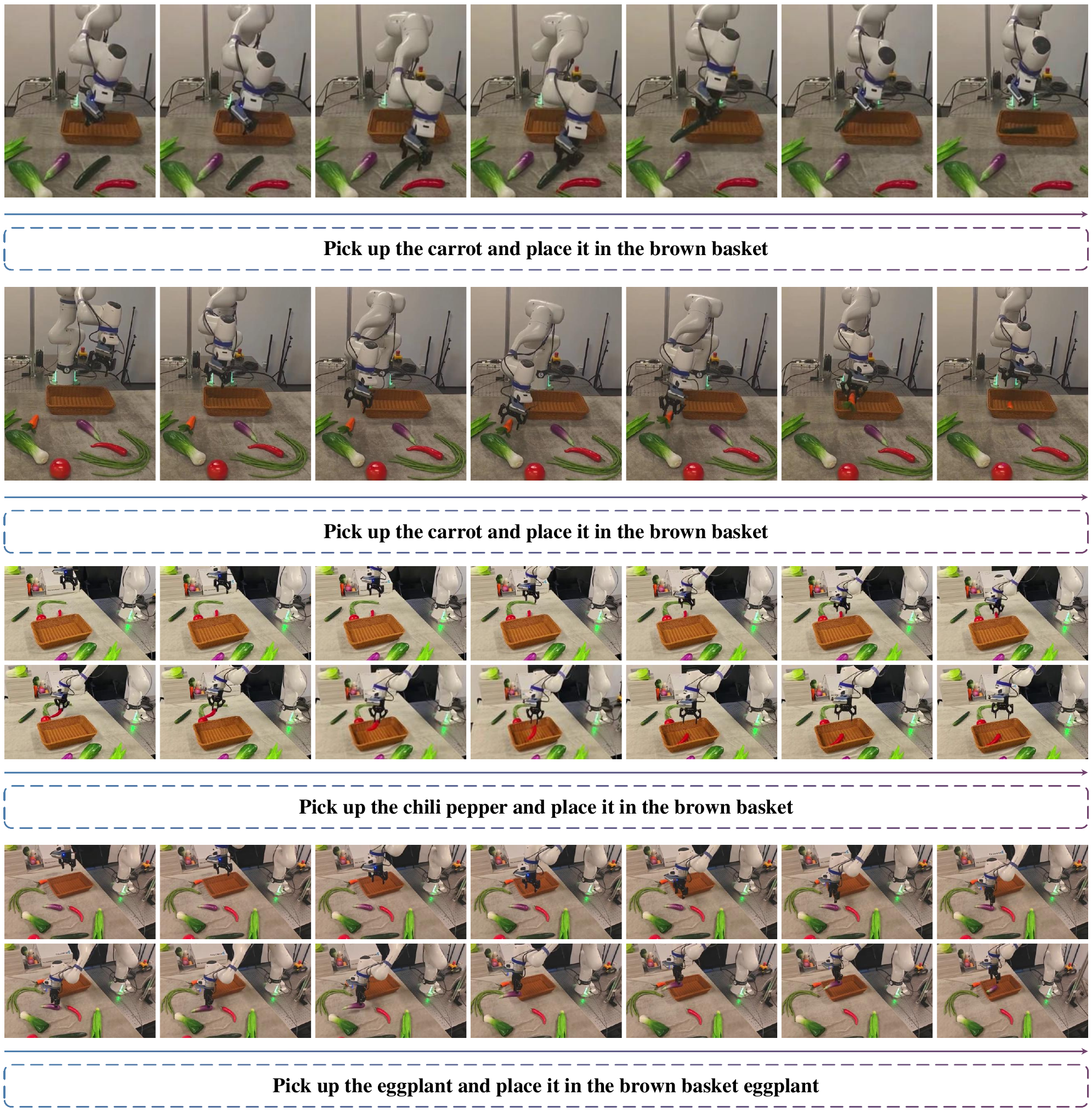}
    \caption{\textbf{Real-world inference examples for vegetable pick-and-place.} We show two rollout sequences where \methodname follows language instructions to pick the specified vegetable and place it into the brown basket.}
    \label{fig:realworld_vis}
\end{figure*}

We evaluate \methodname on a real-world robotic setup using a Franka Research 3 arm equipped with an Intel RealSense D435 camera, as visualized in Figure~\ref{fig:real_world_show}.
To assess the model's instruction-following capabilities and generalization, we design a ``Pick and Place All Blocks'' task where the robot must pick up all specific colored blocks (Red, Blue, Yellow, or Green) in the scene and place them into a box.
For the training set, we collect 100 expert demonstrations for each of the Green, Blue, and Yellow blocks. The Red block is held out to test Out-Of-Distribution (OOD) generalization.
Crucially, during data collection, we ensure that multiple colored blocks are present in the scene simultaneously.
During testing, we further introduce distractors such as cabinets and wrenches into the environment to rigorously evaluate robustness.
 This setup forces the model to rely on language instructions to identify the correct target, rather than exploiting simple visual shortcuts (e.g., reacting to the only object in the scene).
We initialize our model using a subset of the Open X-Embodiment (OXE) dataset, comprising 20 subsets with end-effector (EEF) control configurations. Detailed statistics of the training datasets are provided in Table~\ref{tab:oxe_subsets}.
To align with the action space of the pre-training data, we collect data and control the robot using end-effector (EEF) pose control.
We also provide four qualitative inference examples for the real-world vegetable pick-and-place task in Figure~\ref{fig:realworld_vis}.

\subsection{Preservation of General Capabilities}\label{app:general_capabilities}

It is important to make a nuanced distinction regarding these results: while \methodname effectively preserves the model's normal text-only conversational abilities, its general vision-language conversation skills (involving both image and text inputs) may still experience some degradation following VLA training.
We hypothesize that this is primarily because our training setup requires the vision tower and multimodal projection layers to adapt for control tasks, inevitably shifting visual representations away from the original pre-trained vision-language alignment manifold.
Such specialization is a natural consequence of optimizing an embodied agent for action execution.

\begin{figure*}[h]
    \centering
    \includegraphics[width=1\textwidth]{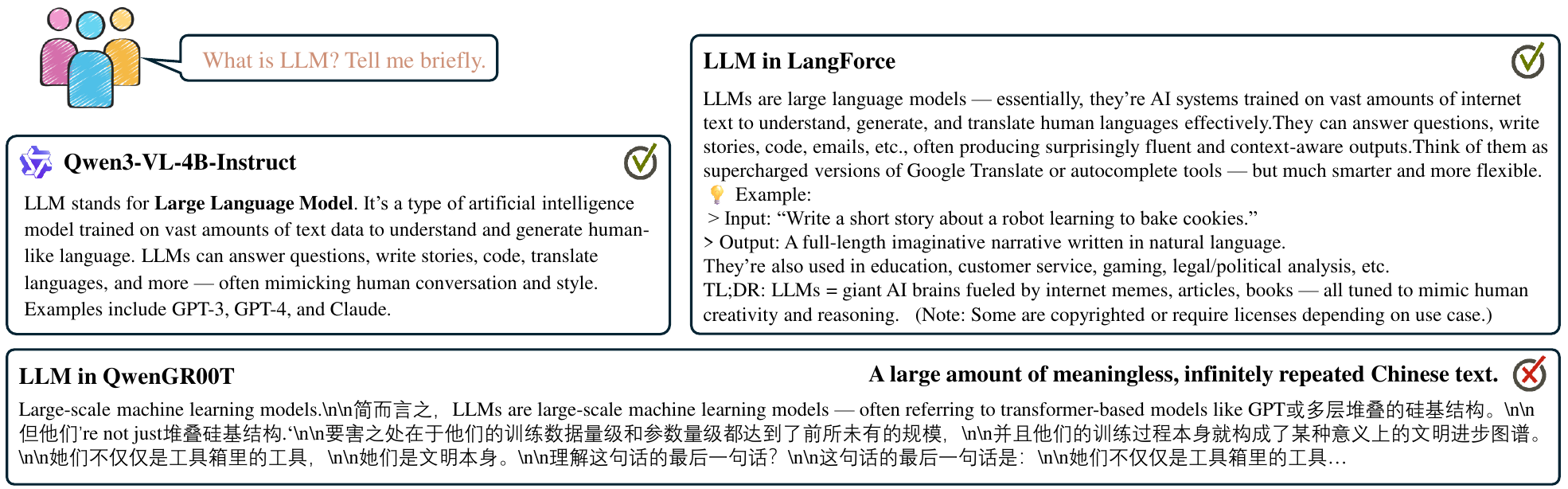}
    \caption{\textbf{Additional qualitative comparison.} Demonstrating the preservation of general VLM capabilities on another example.}
    \label{fig:qa2}
\end{figure*}

Crucially, however, \methodname prevents the collapse of the backbone's \emph{text-only language behavior}.
In standard baselines, the prevalence of vision shortcuts often renders instruction tokens effectively redundant for control. This weakens the training signal that reinforces meaningful language processing, leading to a drift in shared parameters that manifests as failures even on pure text queries (as seen in Fig.~\ref{fig:qa} and Fig.~\ref{fig:qa2}).
Conversely, our method's LLR objective enforces a strong, explicit dependency on language.
This acts as a regularizer, maintaining the functional utility of instruction tokens and thereby preserving the backbone's text-only conversational ability, even as the visual modality specializes for control.
This preservation is of significant practical value, ensuring that the VLM backbone does not degenerate into a shallow feature mapper. 
By retaining its linguistic core, the agent preserves the potential for higher-level reasoning and generalization to novel instructions—key motivations for employing Foundation Models in robotics.

\input{tab/auxiliary_language_ablation}

\begin{figure*}[!t]
    \centering
    \includegraphics[width=0.98\textwidth]{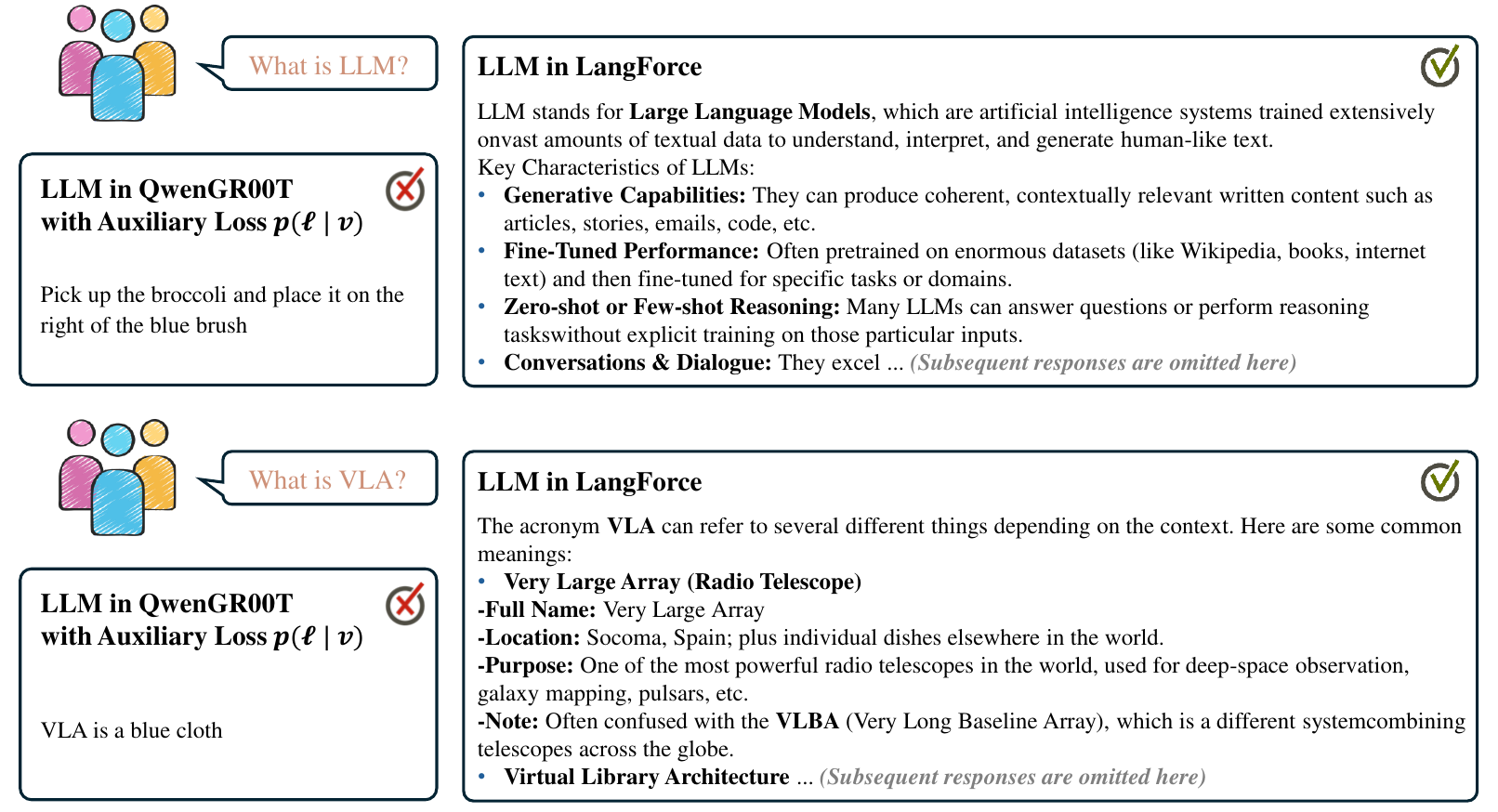}
    \caption{\textbf{Qualitative examples for the auxiliary language-prediction ablation.} Adding an auxiliary $p(\ell \mid v)$ loss helps preserve surface conversational fluency, but the model tends to overfit toward robot-training instructions or scene-specific associations.}
    \label{fig:auxiliary_language_capability}
\end{figure*}

\paragraph{Auxiliary language-prediction ablation.}
To separate generic language retention from action-relevant language grounding, we additionally evaluate a control variant that keeps the same QwenGR00T backbone but adds an auxiliary language-prediction objective for $p(\ell \mid v)$ during VLA fine-tuning.
This is close to adding an extra SFT-style loss that encourages the model to reconstruct the training instruction from the visual scene.
As shown in Table~\ref{tab:auxiliary_language_ablation}, this variant provides only a small policy gain over QwenGR00T (56.9\% vs. 55.2\%) and remains substantially below \methodname{} (66.5\%).
Qualitatively, as shown in Fig.~\ref{fig:auxiliary_language_capability}, the auxiliary-loss model can preserve surface fluency on text-only prompts, but it often overfits to robot-training instructions, producing command-like or scene-specific responses to open-domain questions.
This behavior is expected: optimizing $p(\ell \mid v)$ strengthens scene-to-instruction prediction and can sharpen the vision shortcut.
In contrast, the LLR objective preserves language by keeping instruction tokens causally useful for action generation, binding language to action-relevant distinctions rather than merely asking the model to reproduce likely training instructions.

\subsection{Ablation on $\lambda$ and $\beta$ in Loss Function}\label{app:ablation_lambda_beta}

We analyze the impact of hyperparameters $\lambda$ and $\beta$ based on the results in Table~\ref{tab:ablation_lambda} and Table~\ref{tab:ablation_beta}.
First, regarding the prior loss weight $\lambda$, we observe that even when $\lambda=0$ (where the prior branch is used solely for LLR calculation without explicit action supervision), the model achieves an average success rate of 63.3\%. This represents a substantial improvement over the QwenGR00T baseline (55.2\%), confirming that the LLR objective alone effectively regularizes the policy.
Increasing $\lambda$ to 0.3 yields the optimal performance of 66.5\%, validating that explicitly learning the vision-only prior $p(a|v)$ further aids the decomposition.
The performance remains robust across $\lambda \in [0, 0.5]$, though it drops slightly at higher values.

Second, for the LLR weight $\beta$, setting $\beta=0$ corresponds to training the dual-branch architecture without the mutual information maximization term. This configuration achieves 61.3\%, surpassing the baseline (55.2\%). This suggests that the architectural design, which explicitly separates a vision-only pathway to absorb dataset biases, inherently helps the posterior branch focus on language.
Incorporating the LLR term ($\beta=0.1$) further boosts the success rate to 66.5\%.

\input{tab/ablation_lamda_beta}

\subsection{Ablation on Number of Latent Action Queries}\label{app:ablation_query}
To explore the potential of Latent Action Queries as a out-of-the-box technique, we conduct this ablation solely on the query mechanism, without incorporating our proposed Bayesian decomposition.
We analyse the impact of the number of latent action queries on model performance, as summarized in Table~\ref{tab:ablation_query}.
We observe that increasing the number of queries from 16 to 64 leads to a substantial improvement in success rate.
However, doubling the queries to 128 results in performance saturation without providing additional benefits.
Consequently, we adopt 64 queries as the optimal configuration to balance performance and computational overhead.

\input{tab/ablation_action}

\input{sec/5_discussion}

\begin{figure*}[h]
    \centering
    \includegraphics[width=1\textwidth]{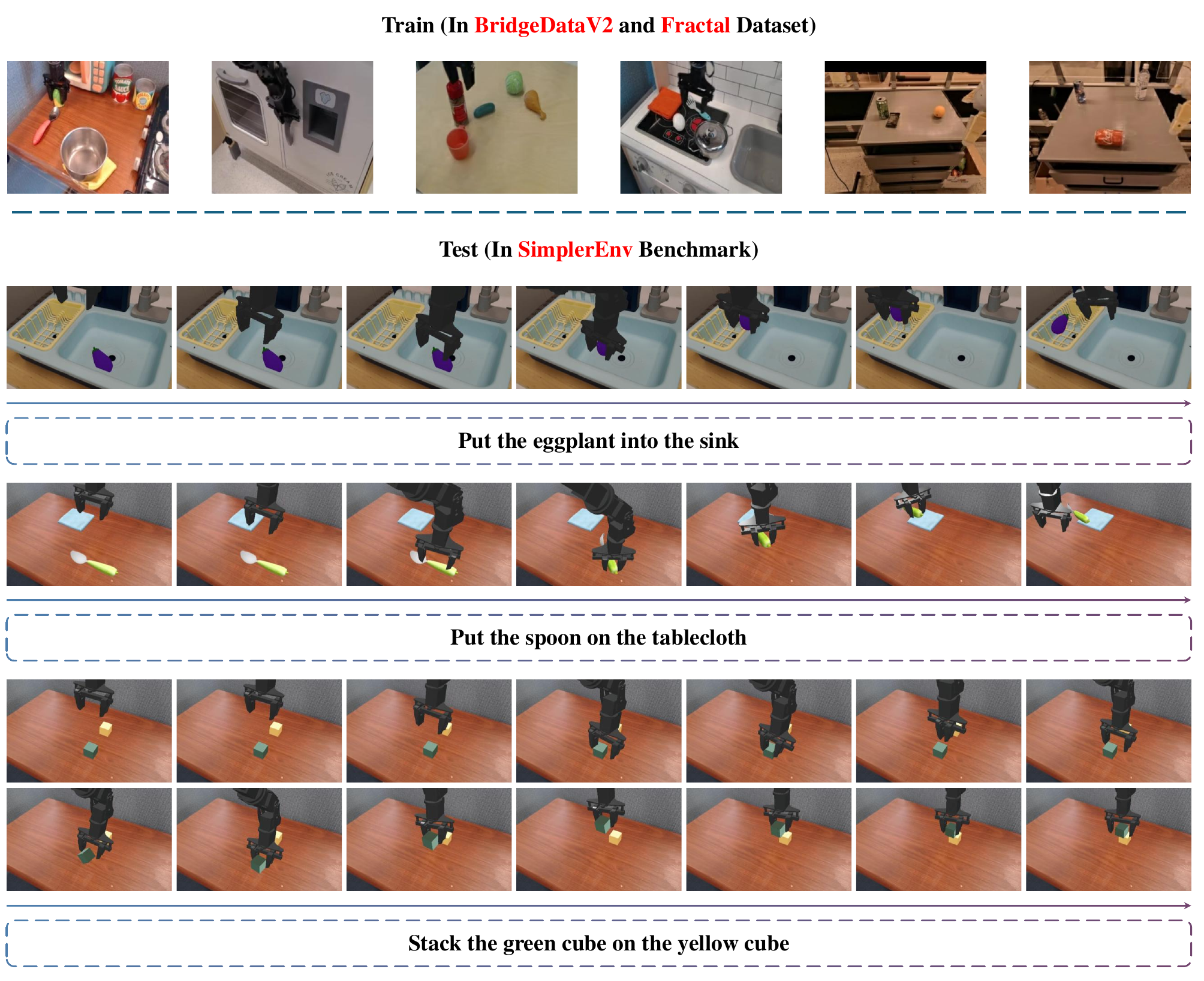}
    \caption{\textbf{Visualization of the domain gap.} Top: Training frames from BridgeDataV2 and Fractal. Bottom: \methodname rollouts in SimplerEnv. More inference videos can be found in the supplementary material.}
    \label{fig:simpler_vis}
\end{figure*}

%% file: tab/libero.tex
\begin{table}[h]
    \centering
    \small
    \caption{\textbf{Comparison on the LIBERO benchmark.} We train one policy for all 4 suites. Avg@500 success rates (\%) across four task suites: Spatial, Object, Goal, and Long.}
    \label{tab:libero_results}
    \setlength{\tabcolsep}{4pt} 
    \begin{tabular}{lccccc}
        \toprule
        \textbf{Method} & \textbf{Spatial} & \textbf{Object} & \textbf{Goal} & \textbf{Long} & \textbf{Avg} \\
        \midrule
        OpenVLA \citep{OpenVLA_2024_CoRL} & 87.4 & 88.4 & 79.2 & 53.7 & 76.5 \\
        OpenVLA-OFT \citep{OpenVLA-OFT_2025_arXiv} & 97.6 & 98.4 & 97.9 & 94.5 & \underline{97.1} \\
        $\pi_0$ \citep{PI0_2024_arXiv} & 96.8 & 98.8 & 95.8 & 85.2 & 94.1 \\
        $\pi_{0.5}$ \citep{PI05_2025_arXiv} & 98.8 & 98.2 & 98.0 & 92.4 & 96.9 \\
        Qwen3-VL-FAST & 97.3 & 97.4 & 96.3 & 90.6 & 95.4 \\
        Qwen3-VL-OFT & 97.8 & 98.6 & 96.2 & 93.8 & 96.6 \\
        Qwen3-VL-GR00T & 97.8 & 98.8 & 97.4 & 92.0 & 96.5 \\
        VisionOnly Qwen3-VL-GR00T & 90.2 & 99.6 & 9.8 & 86.0 & 71.4 \\
        \rowcolor{gray!30}
        \textbf{\methodname{}} & 99.2 & 99.6 & 99.4 & 95.2 & \textbf{98.4} \\
        \bottomrule
    \end{tabular}
\end{table}

%% file: tab/app_realworld.tex
\begin{table}[!h]
    \centering
    \small 
    \caption{\textbf{Selected OXE Subsets for Pre-training.} We curate 20 subsets from the Open X-Embodiment dataset that utilize end-effector position control. The table details the dataset source, the robot platform, and the number of episodes utilized.}
    \label{tab:oxe_subsets}
    \setlength{\tabcolsep}{15pt} 
    \begin{adjustbox}{width=0.96\textwidth}
    \begin{tabular}{llcllc}
        \toprule
        \textbf{Dataset Name} & \textbf{Robot} & \textbf{Episodes} & \textbf{Dataset Name} & \textbf{Robot} & \textbf{Episodes}\\
        \midrule  
        RT-1 Robot Action & Google Robot & 79,499 & BC-Z & Google Robot & 39,350 \\
        Berkeley Bridge & WidowX & 25,460 & Berkeley Fanuc Manipulation & Fanuc Mate  & 415\\ 
        Berkeley Autolab UR5 & UR5 & 896 & UCSD Kitchen & xArm & 150 \\
        USC Jaco Play & Jaco 2 & 976 & Stanford Kuka Multimodal & Kuka iiwa & 3,000 \\
        CMU Stretch & Hello Stretch & 135 & DobbE & Hello Stretch & 5,208 \\
        NYU VINN & Hello Stretch & 435 & - & - & - \\
        \midrule
        DROID & Franka & 92,233 & Austin BUDS & Franka & 50 \\
        Freiburg Franka Play & Franka & 3,242 & CMU Franka Pick-Insert Data & Franka & 520 \\
        Stanford HYDRA & Franka & 550 & Austin Mutex & Franka & 1,500 \\
        Austin VIOLA & Franka & 135 & CMU Play Fusion & Franka & 576 \\
        NYU Franka Play & Franka & 456 & - & - & - \\
        \midrule
        \textbf{Total} & & & & & \textbf{254,786} \\
        \bottomrule
    \end{tabular}
    \end{adjustbox}
\end{table}

%% file: tab/robocasa.tex
\begin{table*}[!t]
    \centering
    \small
    \renewcommand{\arraystretch}{1.3} 
    \setlength{\tabcolsep}{1.6pt} 

    \caption{
      \textbf{Results of evaluating the VLA models with the GR1 robot in the RoboCasa Tabletop simulation environment}. The results for Isaac-GR00T N1.5 and Isaac-GR00T N1.6 are sourced from the official Isaac-GR00T github repository~\citep{GR00T_2025_arXiv}. The results for QwenGR00T, QwenPI, QwenOFT, QwenFAST are sourced from the official starVLA experiments~\citep{starvla_2025}. We highlight the best results in \textbf{bold} and the second-best results with \underline{underline}.
    }
    \begin{tabular}{l c c c c c c c c}
        \toprule
        \rowcolor{white} 
        {Task} & 
        {\scriptsize \makecell{Isaac-GR00T\\N1.5}} & 
        {\scriptsize \makecell{Isaac-GR00T\\N1.6}} & 
        {\scriptsize \makecell{QwenGR00T\\+Qwen3VL}} & 
        {\scriptsize \makecell{QwenPI\\+Qwen3VL}} & 
        {\scriptsize \makecell{QwenOFT\\+Qwen3VL}} & 
        {\scriptsize \makecell{QwenFAST\\+Qwen3VL}} &
        {\scriptsize \makecell{VisionOnly \\QwenGR00T}} & 
        {\scriptsize \makecell{\methodname\\+Qwen3VL}} \\
        \midrule
        PnP Bottle To Cabinet Close                                    & 54.0 & 51.5 & 46.0 & 26.0 & 30.0 & 38.0 & 70.0 & 72.0\\
        PnP Can To Drawer Close                                        & 50.0 & 13.0 & 80.0 & 62.0 & 76.0 & 44.0 & 78.0 & 78.0\\
        PnP Cup To Drawer Close                                        & 38.0 &  8.5 & 54.0 & 42.0 & 44.0 & 56.0 & 42.0 & 46.0\\
        PnP Milk To Microwave Close                                    & 60.0 & 14.0 & 48.0 & 50.0 & 44.0 & 44.0 & 50.0 & 56.0\\
        PnP Potato To Microwave Close                                  & 32.0 & 41.5 & 28.0 & 42.0 & 32.0 & 14.0 & 44.0 & 36.0\\
        PnP Wine To Cabinet Close                                      & 38.0 & 16.5 & 46.0 & 32.0 & 36.0 & 14.0 & 40.0 & 46.0\\
        \midrule
        \rowcolor{gray!20}PnP * to * Close (Avg)                       & 45.3 & 24.2 & 50.3 & 42.3 & 43.7 & 35.0 & 54.0 & 55.7\\
        \midrule
        PnP Novel From Cuttingboard To Basket                          & 38.0 & 58.0 & 48.0 & 40.0 & 50.0 & 54.0 & 58.0 & 66.0\\
        PnP Novel From Cuttingboard To Cardboardbox                    & 46.0 & 46.5 & 40.0 & 46.0 & 40.0 & 42.0 & 26.0 & 40.0\\
        PnP Novel From Cuttingboard To Pan                             & 58.0 & 68.5 & 68.0 & 60.0 & 70.0 & 58.0 & 72.0 & 68.0\\
        PnP Novel From Cuttingboard To Pot                             & 62.0 & 65.0 & 52.0 & 40.0 & 54.0 & 58.0 & 50.0 & 48.0\\
        PnP Novel From Cuttingboard To Tieredbasket                    & 28.0 & 46.5 & 56.0 & 44.0 & 38.0 & 40.0 & 20.0 & 44.0\\
        \midrule
        \rowcolor{gray!20}PnP Novel From Cuttingboard To * (Avg)       & 46.4 & 56.9 & 52.8 & 46.0 & 50.4 & 50.4 & 45.2 & 53.2\\
        \midrule
        PnP Novel From Placemat To Basket                              & 30.0 & 58.5 & 42.0 & 44.0 & 32.0 & 36.0 & 48.0 & 54.0\\
        PnP Novel From Placemat To Bowl                                & 60.0 & 57.5 & 44.0 & 52.0 & 58.0 & 38.0 & 32.0 & 62.0\\
        PnP Novel From Placemat To Plate                               & 56.0 & 63.0 & 48.0 & 50.0 & 52.0 & 42.0 & 34.0 & 52.0\\
        PnP Novel From Placemat To Tieredshelf                         & 36.0 & 28.5 & 18.0 & 28.0 & 24.0 & 18.0 & 16.0 & 24.0\\
        \midrule
        \rowcolor{gray!20}PnP Novel From Placemat To * (Avg)           & 45.5 & 51.9 & 38.0 & 43.5 & 41.5 & 33.5 & 32.5 & 48.0\\
        \midrule
        PnP Novel From Tray To Cardboardbox                            & 52.0 & 51.5 & 38.0 & 34.0 & 44.0 & 28.0 & 50.0 & 50.0\\
        PnP Novel From Tray To Plate                                   & 48.0 & 71.0 & 56.0 & 64.0 & 56.0 & 34.0 & 64.0 & 58.0\\
        PnP Novel From Tray To Pot                                     & 60.0 & 64.5 & 50.0 & 44.0 & 62.0 & 46.0 & 52.0 & 62.0\\
        PnP Novel From Tray To Tieredbasket                            & 52.0 & 57.0 & 36.0 & 50.0 & 54.0 & 36.0 & 42.0 & 44.0\\
        PnP Novel From Tray To Tieredshelf                             & 32.0 & 31.5 & 16.0 & 28.0 & 30.0 & 16.0 & 16.0 & 22.0\\
        \midrule
        \rowcolor{gray!20}PnP Novel From Tray To * (Avg)               & 48.8 & 55.1 & 39.2 & 44.0 & 49.2 & 32.0 & 44.8 & 47.2\\
        \midrule
        PnP Novel From Plate To Bowl                                   & 58.0 & 57.0 & 60.0 & 52.0 & 60.0 & 52.0 & 26.0 & 54.0\\
        PnP Novel From Plate To Cardboardbox                           & 44.0 & 43.5 & 50.0 & 40.0 & 50.0 & 30.0 & 38.0 & 48.0\\
        PnP Novel From Plate To Pan                                    & 60.0 & 51.0 & 54.0 & 36.0 & 66.0 & 48.0 & 44.0 & 54.0\\
        PnP Novel From Plate To Plate                                  & 64.0 & 78.7 & 70.0 & 48.0 & 68.0 & 50.0 & 60.0 & 78.0\\
        \midrule
        \rowcolor{gray!20}PnP Novel From Plate To * (Avg)              & 56.5 & 57.6 & 58.5 & 44.0 & 61.0 & 45.0 & 42.0 & 58.5\\
        \midrule
        \rowcolor{gray!30} 
        {Average}                                                      & 48.2 & 47.6 & 47.8 & 43.9 & \underline{48.8} & 39.0 & 44.7 & \textbf{52.6}\\
        \bottomrule
    \end{tabular}
    \vskip -0.5em
    \label{tab:robocasa_supp_tab}
\end{table*}

%% file: tab/auxiliary_language_ablation.tex
\begin{table}[h]
  \centering
  \caption{
    \textbf{Ablation on auxiliary language prediction.}
    Adding an auxiliary $p(\ell \mid v)$ objective to QwenGR00T provides only limited policy benefit and remains far below \methodname{}, indicating that preserving generic language behavior alone is insufficient unless language is tied to action-relevant distinctions.
  }
  \begin{tabular}{l c c c c c}
    \toprule
    \textbf{Method}
     & \makecell[c]{\textbf{Spoon on Towel}} 
     & \makecell[c]{\textbf{Carrot on Plate}} 
     & \makecell[c]{\textbf{Stack Block}} 
     & \makecell[c]{\textbf{Eggplant in Basket}} 
     & \textbf{Avg} \\
    \midrule
    QwenGR00T & 87.5 & 50.0 & 29.2 & 54.2 & 55.2 \\
    + Aux. $p(\ell \mid v)$ & 79.6 & 61.6 & 16.6 & 69.7 & 56.9 \\
    \textbf{\methodname} & \textbf{89.6} & \textbf{63.8} & \textbf{33.3} & \textbf{79.2} & \textbf{66.5} \\
    \bottomrule
  \end{tabular}
  \vspace{-1em}
  \label{tab:auxiliary_language_ablation}
\end{table}

%% file: tab/ablation_lamda_beta.tex
\definecolor{navyblue}{HTML}{0071BC}

\begin{table}[h]
  \centering
  \renewcommand{\arraystretch}{0.9} 
  \caption{
    \textbf{Ablation study on $\beta$ in Eq. \ref{eq:total_loss}}. All experiments are based on the Qwen3-VL-4B backbone.
  }
  
  \begin{tabular}{l c c c c c}
    \toprule
    \textbf{$\lambda$}
     & \makecell[c]{\textbf{Put Spoon on Towel}} 
     & \makecell[c]{\textbf{Put Carrot on Plate}} 
     & \makecell[c]{\textbf{Stack Block}} 
     & \makecell[c]{\textbf{Eggplant in Basket}} 
     & \textbf{Avg} \\
    \midrule
    0    & 86.4 & 64.6 & 26.0 & 76.0 & 63.3  \\
    0.1  & 84.2 & 65.6 & 27.3 & 81.9 & 64.8 \\
    \textbf{0.3}  & 89.6 & 63.8 & 33.3 & 79.2 & 66.5 \\
    0.5  & 81.2 & 61.5 & 25.8 & 87.7 & 64.1 \\
    \bottomrule
  \end{tabular}
  \label{tab:ablation_lambda}
\end{table}

\begin{table}[h]
  \centering
  \renewcommand{\arraystretch}{0.9} 
  \caption{
    \textbf{Ablation study on $\beta$ in Eq. \ref{eq:total_loss}}. All experiments are based on the Qwen3-VL-4B backbone.
  }
  
  \begin{tabular}{l c c c c c}
    \toprule
    \textbf{$\beta$}
     & \makecell[c]{\textbf{Put Spoon on Towel}} 
     & \makecell[c]{\textbf{Put Carrot on Plate}} 
     & \makecell[c]{\textbf{Stack Block}} 
     & \makecell[c]{\textbf{Eggplant in Basket}} 
     & \textbf{Avg} \\
    \midrule
    0    & 80.8 & 61.0 & 30.8 & 72.5 & 61.3 \\
    \textbf{0.1}  & 89.6 & 63.7 & 33.3 & 79.2 & 66.5 \\
    0.2  & 82.5 & 67.5 & 34.2 & 77.9 & 65.5 \\
    0.3  & 73.8 & 58.3 & 29.2 & 85.2 & 61.6 \\
    \bottomrule
  \end{tabular}
  \label{tab:ablation_beta}
\end{table}

%% file: tab/ablation_action.tex
\definecolor{navyblue}{HTML}{0071BC}

\begin{table}[h]
  \centering
  \renewcommand{\arraystretch}{0.9} 
  \caption{
    \textbf{Ablation study on Number of Latent Action Query}. All experiments are based on the Qwen3-VL-4B backbone.
    }
  
  \begin{tabular}{l c c c c c}
    \toprule
    \textbf{Number}
     & \makecell[c]{\textbf{Put Spoon on Towel}} 
     & \makecell[c]{\textbf{Put Carrot on Plate}} 
     & \makecell[c]{\textbf{Stack Block}} 
     & \makecell[c]{\textbf{Eggplant in Basket}} 
     & \textbf{Avg} \\
    \midrule
    16  & 69.8 & 47.9 & 20.8 & 60.4 & 49.7 \\
    32  & 62.5 & 57.4 & 25.8 & 79.2 & 56.2 \\
    64  & 74.6 & 58.3 & 29.2 & 67.9 & 57.5 \\
    128 & 72.9 & 57.3 & 29.1 & 70.8 & 57.5 \\
    
    \bottomrule
  \end{tabular}
  \label{tab:ablation_query}
\end{table}

%% file: sec/5_discussion.tex
\section{Discussion}\label{sec:discussion}

Based on our analysis of the vision shortcut and the Bayesian decomposition framework, we discuss several potential insights that may guide future research and community practices.

\noindent\textbf{Rethinking Data Collection Strategies.}
Our experiments suggest that the deterministic mapping from visual scenes to language instructions ($H(\ell \mid v) \approx 0$) in goal-driven datasets is a significant factor contributing to the vision shortcut. To mitigate this, we hypothesize that a shift in data collection strategies could be beneficial. Prioritizing data collection in ambiguous scenarios—where the task cannot be inferred solely from the initial observation—might naturally increase the conditional entropy of language. By enriching datasets with scenes that support multiple valid tasks, models may be forced to rely more heavily on instructions for disambiguation.

\noindent\textbf{Leveraging Human Data for Robustness.}
Recently, there has been growing interest in training robot models on large-scale human video data, such as HRDT~\citep{H-RDT_2025_arXiv}, In-N-On~\citep{In-N-On_2025_arXiv}, METIS~\citep{METIS_2025_arXiv}, and PhysBrain~\citep{PhysBrain_2025_arXiv}. Unlike curated robot datasets, human activities are inherently multimodal and context-dependent; the same environment often hosts a wide variety of behaviors, potentially leading to a less sharp $p(\ell \mid v)$. We conjecture that injecting action knowledge from such rich human distributions might help mitigate the information collapse observed in robot-only datasets.

\noindent\textbf{World Models as an Alternative Bayesian Formulation.}
Beyond the VLM framework focused on in this work, recent studies have also explored adapting World Models for VLA control, as seen in F1-VLA~\citep{F1-VLA_2025_arXiv}, Mantis~\citep{Mantis_2025_arXiv}, and InternVLA-A1~\citep{InternVLA-A1_2026_arXiv}.
From a theoretical perspective, these approaches can be viewed as an alternative instantiation of the Bayesian rule, specifically performing inverse dynamics on imagined futures. If we consider $v$ as a sequence of past frames $v_{\le t}$, and treat the future state $v_{t+1}$ as a latent variable generated by the model (conditioned on $\ell$), the action inference can be expressed as:
\begin{equation}
    p(a \mid v_{\le t}, v_{t+1}, \ell) = \frac{p(v_{t+1} \mid v_{\le t}, a, \ell) \, p(a \mid v_{\le t}, \ell)}{p(v_{t+1} \mid v_{\le t}, \ell)}.
\end{equation}
Here, the numerator $p(v_{t+1} \mid v_{\le t}, a, \ell)$ represents a \textit{world model} (forward dynamics) predicting the future state. The term $p(a \mid v_{\le t}, \ell)$ serves as an action prior, and the denominator $p(v_{t+1} \mid v_{\le t}, \ell)$ represents the future prediction marginalized over actions.
In this formulation, the policy execution involves first ``imagining'' a desired future $v_{t+1}$ consistent with $\ell$, and then inferring the optimal action $a$ via the equation above.
Since world models are typically trained on vast amounts of video data, the predictive distribution (the numerator) is often rich and highly sensitive to the action $a$. We hypothesize that this sensitivity prevents the collapse of the numerator to the denominator. This suggests that world model-based architectures could offer another robust technical path toward solving the vision shortcut, which we plan to explore in future work.